%% file: neurips_2026.tex
\documentclass{article}

\PassOptionsToPackage{numbers, compress}{natbib}
 \usepackage[eandd, final]{neurips_2026}

\usepackage[utf8]{inputenc} % allow utf-8 input
\usepackage[T1]{fontenc}    % use 8-bit T1 fonts
\usepackage{hyperref}       % hyperlinks
\usepackage{url}            % simple URL typesetting
\usepackage{booktabs}       % professional-quality tables
\usepackage{amsfonts}       % blackboard math symbols
\usepackage{nicefrac}       % compact symbols for 1/2, etc.
\usepackage{microtype}      % microtypography
\usepackage{xcolor}         % colors
\usepackage{xspace}
\usepackage{multirow}
\usepackage{multicol}
\usepackage{graphicx}
\usepackage{tikz}
\usepackage{pgfplots}
\usepackage{wrapfig}
\usepackage{titletoc}
\usepackage{enumitem}
\usepackage{subcaption}
\usepackage{tabularx}
\usepackage{array}
\usepackage{pifont}
\usepackage{todonotes}
\usepackage{tcolorbox}
\usepackage{amsmath} 

\pgfplotsset{compat=1.18} % 버전에 맞게 조정

\newcommand{\ourdataset}{NarraScene\xspace}
\newcommand{\ourbaseline}{Narrative-Prompted VLM\xspace}

\newcommand{\eg}{\textit{e.g.}}

\title{Beyond Visual Boundaries: Rethinking \\ Scene Segmentation for Movie RAG}
\author{%
  \textbf{Dong-Hee Kim\textsuperscript{1}} \quad
  \textbf{Seonwoo Choi\textsuperscript{1}} \quad
  \textbf{Changbeen Kim\textsuperscript{1}} \quad
  \textbf{Jungmyung Wi\textsuperscript{1}}
  \\[3pt]
  \textbf{Juyeon Ko\textsuperscript{2}} \quad
  \textbf{Youngju Choi\textsuperscript{2}} \quad
  \textbf{Il Hyeon Mun\textsuperscript{2}} \quad
  \textbf{Hyunwoo J. Kim\textsuperscript{3}} \quad
  \textbf{Donghyun Kim\textsuperscript{1}}
  \\[5pt]
  \normalfont
  \textsuperscript{1}Korea University
  \qquad
  \textsuperscript{2}KT Corporation
  \qquad
  \textsuperscript{3}KAIST
}

\begin{document}

\maketitle

\input{00_abstract}

\input{01_introduction}

\input{02_related}

\input{03_our_dataset}

\input{04_baseline}

\input{05_experiments}

\input{06_conclusion}

\begin{ack}
This work was the result of a project supported by the KT (Korea Telecom)--Korea University AICT R\&D Center (60\%). 
% Juyeon Ko, Youngju Choi, and Il Hyeon Mun are employees of KT Corporation  
This research was supported by the Institute of Information \&communications Technology Planning \& Evaluation (IITP) grant funded by the Korea government(MSIT) (No. RS-2019-II190079, Artificial Intelligence Graduate School Program(Korea University), 1\%) and the National Research Foundation of Korea(NRF) grant funded by the Korea government(MSIT)(No. RS-2024-00341514, 39\%).
The authors declare no other competing interests.
\end{ack}

\bibliographystyle{abbrvnat}   % 또는 plainnat, unsrtnat 등
\bibliography{references}      % references.bib 파일명에서 .bib 확장자 제거

%%%%%%%%%%%%%%%%%%%%%%%%%%%%%%%%%%%%%%%%%%%%%%%%%%%%%%%%%%%%
\newpage

\appendix

\input{99_appendix}

%%%%%%%%%%%%%%%%%%%%%%%%%%%%%%%%%%%%%%%%%%%%%%%%%%%%%%%%%%%%

% \newpage
% \input{checklist.tex}

\end{document}

%% file: 00_abstract.tex
\begin{abstract}
    Understanding long-form video remains a fundamental challenge for multimodal large language models (MLLMs). Sparse frame sampling fails to capture fine-grained visual details, while dense sampling quickly exceeds context length limits. Retrieval-augmented generation (RAG) offers a promising middle ground by selectively retrieving relevant video segments for grounded generation, yet its effectiveness critically depends on the quality of the video segments used as retrieval units.
	In this paper, we investigate RAG for movie understanding, which demands story-level reasoning over characters, events, and narrative arcs spanning hours of content. Scene segmentation, a long-studied problem that partitions movies into semantically coherent units, is a natural candidate for defining such retrieval units. We reexamine whether existing methods actually serve this role through comprehensive evaluation on downstream movie understanding tasks, and find that they do not provide a consistent or dominant advantage over naive uniform temporal chunking. Our audit of the most standard scene segmentation benchmarks reveals why: current annotations prioritize visually salient transitions over narrative event structure. 
	Motivated by this mismatch, we introduce \ourdataset, a narrative-centric scene segmentation dataset annotated with a three-level cognitive taxonomy spanning physical, character, and narrative change, where every valid boundary requires a narrative-level shift. When used as retrieval units, these narrative-grounded segments outperform uniform chunking on downstream movie understanding tasks, suggesting that the central challenge for scene segmentation in movie RAG is not detecting boundaries, but identifying the narrative event units that matter for movie understanding. The dataset is available at \url{https://huggingface.co/datasets/queez0405/NarraScene}.
    
    % Scene segmentation is widely used as a chunking abstraction for long-form movie understanding, especially in modern retrieval-augmented Multimodal Large Language Model (MLLM) pipelines that must organize and revisit evidence from hour-scale videos. In this work, we reexamine whether existing scene segmentation methods actually provide useful context units for movie understanding tasks. Across movie temporal retrieval, question answering, and fact verification, we find that scene segmentation methods trained on legacy dataset consistently fail to outperform simple uniform splitting. To understand this transfer gap, we audit MovieNet scene annotations under an Event Segmentation Theory based definition of scene boundaries and find that existing annotations are more strongly aligned with visually salient transitions than with narrative event change. Motivated by this mismatch, we introduce \ourdataset, a narrative-centric scene segmentation dataset built on full-length movies and annotated with a three-level cognitive taxonomy over physical change, character change, and narrative change, where every valid boundary must include a narrative-level shift. Finally, we show that when these hand-labeled narrative scenes are used as retrieval units, they outperform uniform chunking on downstream movie claim verification. Our results suggest that the central challenge for scene segmentation in movie RAG is not only detecting boundaries, but identifying the narrative event units that matter for movie understanding.
\end{abstract}

%% file: 01_introduction.tex
\vspace{-2mm}
\section{Introduction}
\vspace{-2mm}

\begin{figure}[t]
  \centering
  \includegraphics[width=\textwidth,trim=2.5cm 0cm 0cm 0cm,clip]{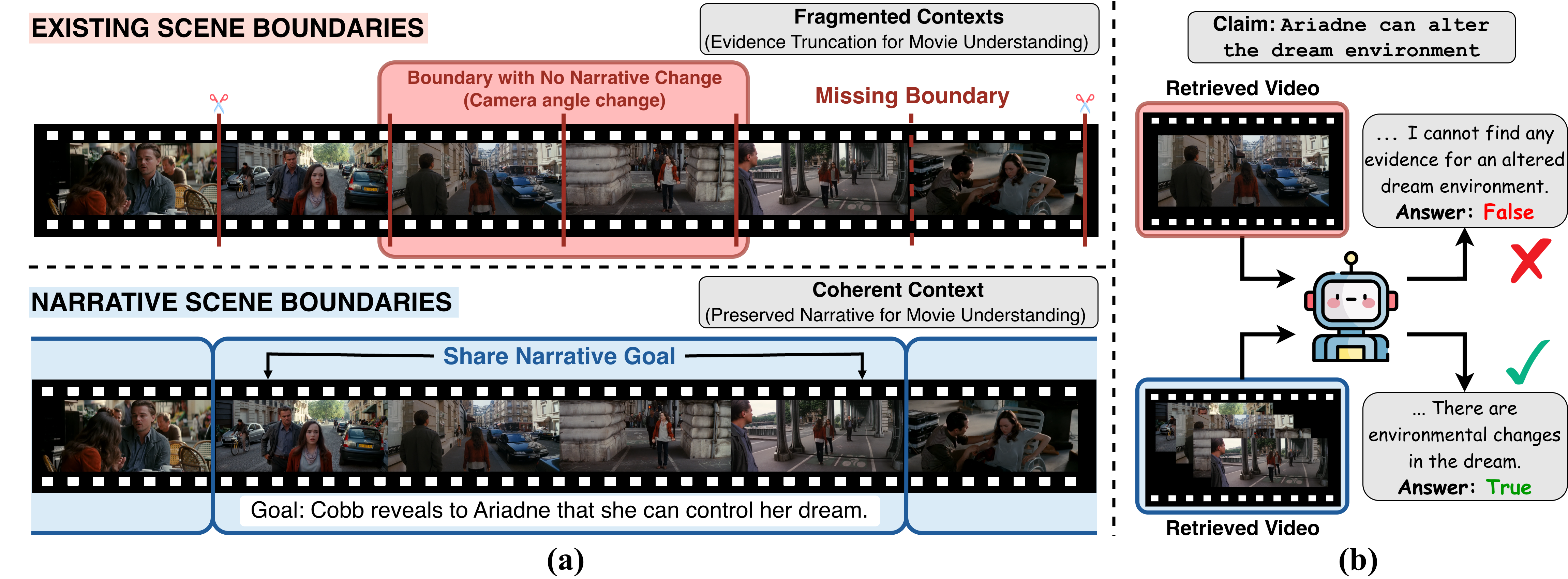}
  \vspace{-15pt}
  \caption{\textbf{Existing scene boundaries fail to support movie RAG.}
  (a) Scene boundary comparison: Standard scene segmentation boundaries often follow visually salient transitions, which can create spurious cuts and miss valid boundaries, fragmenting a continuous narrative event. In contrast, our definition groups shots by shared narrative goal and preserves coherent context.
  (b) Movie RAG example: This difference directly affects retrieval-based movie claim verification. Fragmented visual chunks can miss relevant evidence and lead to an incorrect answer, while narrative scene boundaries retrieve a coherent event for movie understanding.}
  \label{fig:introduction}
  \vspace{-5pt}
\end{figure}

Despite recent advances in multimodal large language models (MLLMs)~\citep{bai2025qwen3vl, wang2025internvl3_5}, long-form video understanding remains a fundamental challenge~\citep{chandrasegaran2024hourvideo, zaranis2025mf2}. One common approach is to sparsely sample frames across the entire video and pass them directly to MLLMs within their context limit~\citep{hu2025mllm,li2025improving}. However, sparse sampling inevitably sacrifices fine-grained visual and temporal details. Naively increasing the number of input frames does not resolve this issue, as current MLLMs struggle to effectively attend to and reason over long sequences of visual tokens~\citep{shu2025videoxl,wei2025visualextension}.

Retrieval-augmented generation (RAG) addresses this limitation by retrieving only the video segments most relevant to a given query and passing them as context to the downstream model~\citep{gia2025vrag}.  Rather than processing the entire video, the model reasons over a small set of targeted evidence segments, effectively decoupling scalability from reasoning. Yet this pipeline is only as good as its retrieval units. When segmentation boundaries fragment a coherent narrative exchange or lump together visually similar but semantically unrelated shots, the downstream model is deprived of the evidence it needs, regardless of how capable the downstream model may be~\citep{zeng2025scenerag,gia2025vrag}. The segmentation strategy is therefore not a peripheral design choice but a foundational one, as it determines what evidence the system can ever retrieve and thereby sets a hard ceiling on downstream performance.

In this paper, we investigate RAG for movie understanding, a domain that demands story-level reasoning over characters, events, and narrative arcs spanning hours of content. Scene segmentation~\citep{mun2022bassl,rao2020lgss,cho2026iclrsceneseg, islam2023trans4mer, chen2021shotcol, wu2022scrl}, which partitions movies into semantically coherent units, is a natural candidate for defining retrieval units in such a pipeline, and has been studied extensively in the computer vision community. However, existing methods have been developed and evaluated primarily on the most standard benchmark MovieNet~\citep{huang2020movienet}, leaving it unclear whether the resulting boundaries align with the retrieval units genuinely required for movie RAG. Addressing this gap is the central motivation of our work.

To directly assess whether existing scene segmentation methods provide effective retrieval units for movie RAG, we conduct controlled experiments across three downstream tasks for movie understanding: temporal grounding~\citep{soldan2022mad}, multi-choice question answering~\citep{he2024storyteller}, and claim verification~\citep{zaranis2025mf2}. Fixing the retrieval backbone and downstream models across all tasks, we vary only the segmentation method and find that models trained on the most widely used scene segmentation dataset~\cite{huang2020movienet} do not provide a consistent or dominant advantage over naive temporal uniform splitting.  This result suggests that standard scene segmentation targets may not align with the context structure needed by downstream movie understanding tasks, utilizing movie RAG.

To understand the root cause of this failure, we closely examine existing scene boundary annotations. In narrative film, scene continuity often persists across shot changes and spatial shifts as long as the underlying goal, interaction, or conversational focus remains coherent~\citep{zacks2007eventsegmentation,magliano2011eventsegfilm}. We hypothesize that existing annotations may not adequately reflect this narrative continuity, instead capturing more superficial visual transitions. To test this hypothesis, we re-annotate a stratified sample of ten films from the MovieNet test set using a scene boundary definition grounded in Event Segmentation Theory~\cite{zacks2007eventsegmentation,magliano2011eventsegfilm}, which characterizes event perception as the segmentation of continuous experience into meaningful units when the underlying situation model changes. Comparing these re-annotations against the original MovieNet labels, we find that the majority of existing boundaries are driven primarily by physical transitions, with only a small fraction aligning with narrative event change under our definition. This suggests that current scene segmentation practice is more strongly aligned with visually salient cues than with narrative event structure. Figure~\ref{fig:introduction} illustrates a representative example of this mismatch: a visually salient transition fragments a single coherent narrative event, whereas a narrative boundary preserves the full event as a coherent unit for downstream reasoning.

Motivated by this mismatch, we build \ourdataset, a narrative-centric scene segmentation dataset on full-length, full-audio movies. Our annotation framework is grounded in Event Segmentation Theory~\citep{zacks2007eventsegmentation,kurby2008segmentation}, which views event perception as the process of segmenting continuous activity into meaningful units when the underlying situation model changes. Concretely, we annotate scene transitions along three levels, namely physical change, character change, and narrative change, and require every valid boundary to include a narrative shift. This makes \ourdataset explicitly target narrative event boundaries rather than visual discontinuities alone.

We instantiate this definition by annotating 53 full-length open-licensed movies from MF$^2$~\citep{zaranis2025mf2} through a three-stage protocol of independent proposal, structured cross-verification, and final adjudication. The resulting benchmark contains 2,371 annotated narrative scenes. Beyond providing new labels, \ourdataset serves as an oracle evaluation target for testing whether retrieval units that preserve narrative events improve movie RAG. Under the same retrieval and downstream verification pipeline, hand-labeled narrative scenes consistently outperform uniform chunking on MF$^2$ claim verification. These results suggest that the central challenge for scene segmentation in movie RAG is not simply to detect more accurate boundaries under legacy benchmarks, but to recover the narrative evidence units that downstream tasks need.

\vspace{-1mm}
Our contributions are as follows.
\vspace{-2mm}
\begin{itemize}[leftmargin=1.5em]
    \item We reexamine scene segmentation as a chunking interface for modern movie RAG and show that existing scene segmentation methods do not provide a reliable advantage over simple uniform chunking across temporal grounding, question answering, and fact verification.
    \item We audit MovieNet scene segmentation annotations under an Event Segmentation Theory based definition and show that existing scene segmentation targets are more strongly aligned with physical boundary cues than with narrative event change.
    \item We introduce \ourdataset, a narrative-centric scene segmentation dataset for full-length movies, in which every valid boundary includes a narrative transition under a three-level boundary definition.
    \item We show that hand-labeled narrative scene boundaries can outperform other chunking methods on downstream movie claim verification, establishing that narrative-valid segmentation is useful for movie RAG in principle.
\end{itemize}

%% file: 02_related.tex
% \section{Related Works}

% MovieNet~\citep{huang2020movienet}. MovieChat-SceneSeg~\citep{cho2026iclrsceneseg} by MovieChat-1k~\citep{song2024moviechat}.

% Scene Segmentation Method BaSSL~\citep{mun2022bassl}, Trans4mer~\citep{islam2023trans4mer}, Scene-VLM~\citep{berman2025scenevlm}, Timesformer~\citep{bertasius21timesformer}, ShotCoL~\citep{chen2021shotcol}, MEGA~\citep{sadoughi2023mega}, ChapterLLaMA~\citep{ventura2025chapterllama}, GenreDur~\citep{cho2026iclrsceneseg}

\vspace{-3mm}
\section{When Scene Segmentation Meets Modern Movie RAG}
\vspace{-3mm}
\label{sec:modern_movie_rag}

Movie RAG pipelines must partition full-length movies into retrievable context units that serve as evidence for downstream task reasoning. Existing scene segmentation methods have claimed to produce semantically coherent chunks that benefit long video understanding and diverse downstream tasks~\citep{bain2020condensed,bain2021frozenintime, chandrasegaran2024hourvideo,song2024moviechat}. We directly test this claim by evaluating whether existing segmentation methods actually improve performance across diverse movie understanding tasks when used as retrieval units. We first describe the experimental setup in Sec.~\ref{sec:exp_setup}.

% Modern movie RAG must partition a full-length movie into retrievable context units before retrieval and reasoning. Prior work often treats scene segmentation as the natural unit for this purpose, but whether existing scene boundaries actually provide better retrieval units than simple fixed-length chunks remains unclear. In this section, we test that assumption directly by fixing the retrieval and reasoning pipeline and varying only the segmentation used to partition each movie.

Then, we evaluate segmented scenes as a retrieval unit on three downstream movie understanding tasks that require evidence selection from full-length movies. These tasks are movie fact verification on MF$^2$ \citep{zaranis2025mf2}, temporal grounding on MAD \citep{soldan2022mad}, and multi-choice question answering on MovieStory101 \citep{he2024storyteller}. We present these downstream evaluation results in Sec.~\ref{sec:down_eval}.

\begin{figure}[t]
    \centering
    \includegraphics[width=\columnwidth]{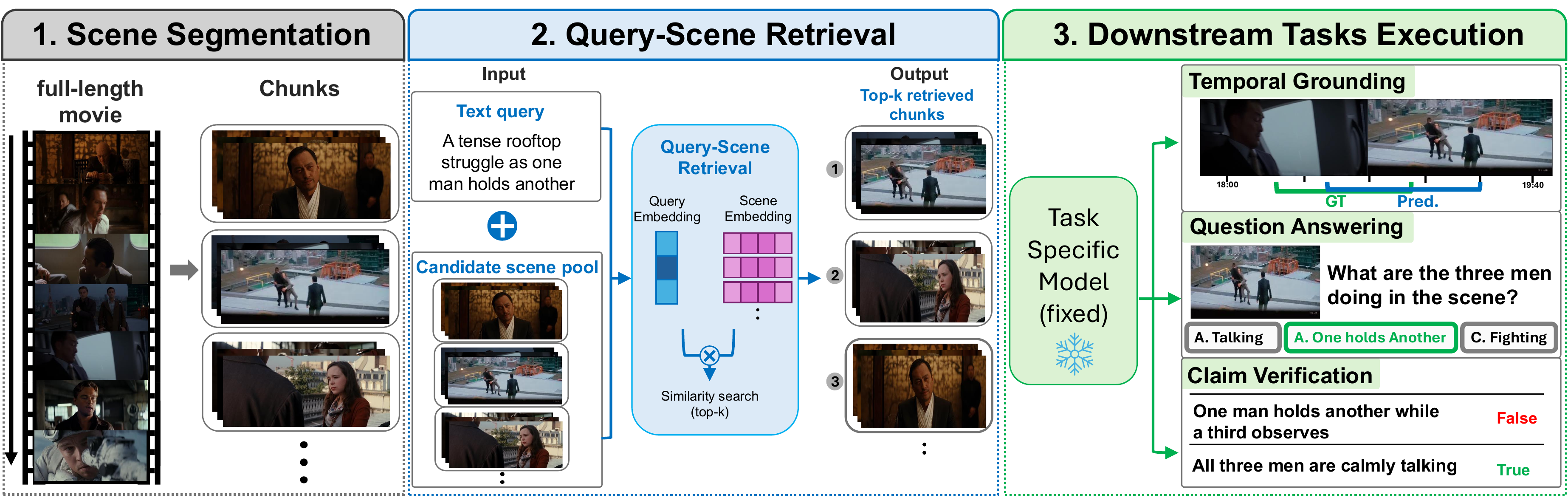}
    \vspace{-18pt}
    \caption{Movie RAG pipeline with scene segmentation-based retrieval. A movie is first partitioned into chunks (segments), segmentation representations are retrieved against a text query, and the retrieved segment is served as evidence for a downstream task.}
    \label{fig:downstream_task_pipeline}
    \vspace{-15pt}
\end{figure}

\subsection{Experimental Setup}
\vspace{-5pt}
\label{sec:exp_setup}
\paragraph{Movie Retrieval Augmented Generation (RAG) Pipeline.}
Given a full-length movie and a downstream movie understanding query (\eg, movie question-answering), we define a movie RAG pipeline consisting of three steps: scene segmentation, query-scene retrieval, and downstream task execution, where retrieved scenes serve as evidence for task-specific models, as illustrated in Figure~\ref{fig:downstream_task_pipeline}.
% To ensure that performance differences stem from the segmentation itself rather than from model architecture,

\vspace{-1mm}
\begin{enumerate}[leftmargin=1.5em, itemsep=0.0em, topsep=0pt, partopsep=0pt, parsep=0pt]
    \item \textbf{Scene Segmentation.} A full-length movie is partitioned into chunks (segments) that reflect coherent narrative units under each segmentation condition.
    \item \textbf{Query-Scene Retrieval.} Given a task-specific text query, we use a cross-modal retrieval model (\eg, \texttt{Qwen3-VL-Embedding-2B}~\citep{qwen3vlembedding}) to retrieve the top-K most relevant chunks.
    \item \textbf{Downstream Tasks Execution.} The retrieved chunks, rather than the full movie, are provided as the sole visual context to a fixed task-specific model (\eg, for movie question answering) to produce the final output.
\end{enumerate}
% \vspace{-1mm}

\noindent\textbf{Scene Segmentation Baselines.} 
To investigate which segmentation approaches yield the most useful retrieval units for movie RAG, we evaluate a diverse set of segmentation methods while fixing the retrieval and downstream models, isolating segmentation quality as the sole variable of interest.
% The primary independent variable in all experiments is the segmentation setting. To separate the effect of boundary placement from the trivial effect of chunk length, we compare four segmentation settings. 
\begin{itemize}[leftmargin=1.5em, itemsep=0.0em, topsep=0pt]
    \item \textbf{Uniform split (chunking)} partitions movies into fixed-length temporal windows without relying on any learned segmentation model.
    \item \textbf{Prior scene segmentation approaches} use scene boundary predictions generated by existing scene segmentation methods. Rather than relying on a single model family, we evaluate several representative approaches, including BaSSL~\citep{mun2022bassl}, GenreDur~\citep{cho2026iclrsceneseg}, and PySceneDetect~\citep{castellano2024pyscenedetect}.
    % \item \textbf{Prior scene segmentation approaches} use scene segmentation predictions from existing scene segmentation methods. For the visual scene boundary setting, we do not rely on a single model family. Instead, we instantiate this setting with representative prior methods including BaSSL \citep{mun2022bassl}, GenreDur \citep{cho2026iclrsceneseg}, and PySceneDetect~\citep{castellano2024pyscenedetect}. 
    \item \textbf{Length-Controlled} baseline preserves the chunk-length statistics of scene-based segmentation while removing the effect of the original scene boundaries. Since existing scene segmentation models often over-segment movies into many short segments, we merge adjacent predicted scenes to form chunks with similar length distributions. This control allows us to isolate whether downstream performance comes from meaningful scene boundaries or from chunk length alone.
\end{itemize}
The experimental design is motivated by a simple but direct question: Do existing scene segmentation methods produce retrieval units that are actually useful for movie RAG? If so, their predicted boundaries should consistently outperform fixed-length uniform chunks across downstream movie understanding tasks. 

\subsection{Downstream Task Evaluation}
\label{sec:down_eval}
We evaluate three downstream movie understanding tasks: temporal grounding, question answering, and claim verification. Below, we briefly introduce the experimental design of each task, with full details provided in Appendix~\ref{app:downstream_task_details}.

\paragraph{Movie Temporal Grounding.}

\begin{table*}[t]
    \centering
    \scriptsize
    \setlength{\tabcolsep}{3.5pt}
    \resizebox{\textwidth}{!}{%
        \begin{tabular}{lccccccc ccccccc}
            \toprule
            \multirow{2}{*}{\textbf{Condition}}
            & \multicolumn{7}{c}{\textbf{TimeLens-8B~\citep{zhang2025timelens}}}
            & \multicolumn{7}{c}{\textbf{Vidi-7B~\citep{team2025vidi}}} \\
            \cmidrule(lr){2-8} \cmidrule(lr){9-15}
            & \textbf{R1@0.1} & \textbf{R5@0.1} & \textbf{R1@0.3} & \textbf{R5@0.3} & \textbf{R1@0.5} & \textbf{R5@0.5} & \textbf{Avg.}
            & \textbf{R1@0.1} & \textbf{R5@0.1} & \textbf{R1@0.3} & \textbf{R5@0.3} & \textbf{R1@0.5} & \textbf{R5@0.5} & \textbf{Avg.} \\
            \midrule
            % \multicolumn{15}{l}{\textit{Whole-Movie Input}} \\
            % \midrule
            Full-length Movie
            & 0.11 & - & 0.00 & - & 0.00 & - & 0.03
            & 2.01 & - & 1.20 & - & 0.62 & - & 1.27 \\
            \midrule
            \multicolumn{15}{l}{\textit{Fixed-Length Retrieval Chunks}} \\
            \midrule
            Uniform Split (2 min)
            & \textbf{12.12} & \textbf{27.42} & \textbf{7.80} & \underline{17.59} & \textbf{4.25} & \underline{9.45} & \textbf{13.10}
            & \underline{7.14} & \textbf{20.48} & \textbf{5.82} & \textbf{12.75} & \textbf{3.28} & \textbf{8.10} & \textbf{9.59} \\
            Uniform Split (3 min)
            & \underline{11.00} & \underline{26.56} & \underline{7.27} & \textbf{17.82} & \underline{4.11} & \textbf{10.36} & \underline{12.85}
            & 5.31 & \underline{17.59} & \underline{3.38} & \underline{10.91} & 1.48 & \underline{7.36} & \underline{7.67} \\
            Uniform Split (4 min)
            & 10.93 & 26.28 & 6.70 & 16.19 & 3.59 & 8.65 & 12.05
            & 5.28 & 15.26 & 2.91 & 10.52 & 1.42 & 6.15 & 6.92 \\
            \midrule
            \multicolumn{15}{l}{\textit{Scene-Based Retrieval Chunks}} \\
            \midrule
            PySceneDetect~\citep{castellano2024pyscenedetect}
            & 3.59 & 9.13 & 1.83 & 7.06 & 1.04 & 1.81 & 4.07
            & 2.28 & 6.71 & 1.19 & 5.32 & 0.79 & 1.25 & 2.92 \\
            BaSSL~\citep{mun2022bassl}
            & 5.36 & 12.14 & 3.10 & 7.90 & 1.69 & 3.67 & 5.64
            & 3.34 & 7.10 & 1.22 & 5.53 & 0.90 & 2.63 & 3.45 \\
            GenreDur~\citep{cho2026iclrsceneseg}
            & 8.19 & 16.96 & 5.09 & 10.10 & 2.56 & 4.77 & 7.94
            & 6.39 & 13.12 & 3.25 & 9.03 & 1.03 & 3.03 & 5.97 \\
            Length-Controlled
            & 8.19 & 21.75 & 3.38 & 11.29 & 2.25 & 6.49 & 8.89
            & \textbf{7.37} & 15.29 & 2.11 & 10.12 & \underline{2.01} & 5.23 & 7.02 \\
            \bottomrule
        \end{tabular}%
    }
    \caption{Comparison of scene segmentation baselines on the temporal grounding benchmark MAD~\citep{soldan2022mad}. We evaluate two recent temporal grounding models (TimeLens, Vidi) under the same movie RAG pipeline. Fixed-length uniform chunks provide strong baselines, while legacy scene segmentation methods consistently underperform.}
    \label{tab:downstream_mad}
    \vspace{-3pt}
\end{table*}

We first investigate this question on MAD ~\citep{soldan2022mad}, a movie temporal grounding benchmark where short target moments must be localized in full movies. This setting is sensitive to chunk boundaries because retrieval succeeds only when the relevant moment is contained in a retrievable unit. Table~\ref{tab:downstream_mad} reports results with two recent grounding models, TimeLens ~\citep{zhang2025timelens} and Vidi~\citep{team2025vidi}. Full-length movie input performs poorly, confirming the need for retrieval-based chunking. Across both models, simple uniform splitting provides the strongest overall baselines, while legacy scene segmentation methods remain below the best uniform setting. GenreDur is the strongest legacy scene baseline, and Length-Controlled is competitive on some Vidi metrics, but neither surpasses the best uniform chunking on average. Overall, standard scene segmentation models do not automatically yield better retrieval units for downstream movie temporal grounding.

\begin{table*}[t]
    \centering
    \scriptsize
    \setlength{\tabcolsep}{3pt}
    \resizebox{\textwidth}{!}{
        \begin{tabular}{l c ccc ccc ccc ccc}
            \toprule
            \multirow{3}{*}{\textbf{Condition}} & \multirow{3}{*}{\textbf{Avg. Length (s)}} & \multicolumn{3}{c}{\textbf{Qwen3-VL (4B)}~\citep{bai2025qwen3vl}} & \multicolumn{3}{c}{\textbf{Qwen3-VL (8B)}~\citep{bai2025qwen3vl}} & \multicolumn{3}{c}{\textbf{InternVL3.5 8B}~\citep{wang2025internvl3_5}} & \multicolumn{3}{c}{\textbf{InternVL3.5 14B}~\citep{wang2025internvl3_5}} \\
            \cmidrule(lr){3-5} \cmidrule(lr){6-8} \cmidrule(lr){9-11} \cmidrule(lr){12-14}
            & & \multicolumn{3}{c}{\textbf{QA Acc. (\%)}} & \multicolumn{3}{c}{\textbf{QA Acc. (\%)}} & \multicolumn{3}{c}{\textbf{QA Acc. (\%)}} & \multicolumn{3}{c}{\textbf{QA Acc. (\%)}} \\
            \cmidrule(lr){3-5} \cmidrule(lr){6-8} \cmidrule(lr){9-11} \cmidrule(lr){12-14}
            & & $K$=1 & $K$=3 & $K$=5 & $K$=1 & $K$=3 & $K$=5 & $K$=1 & $K$=3 & $K$=5 & $K$=1 & $K$=3 & $K$=5 \\
            \midrule
            % \multicolumn{14}{l}{\textit{Whole-Movie Input}} \\
            % \midrule
            Full-length Movie
            & -- 
            & \multicolumn{3}{c}{47.2} 
            & \multicolumn{3}{c}{48.6}
            & \multicolumn{3}{c}{51.9}
            & \multicolumn{3}{c}{55.2} \\
            \midrule
            \multicolumn{14}{l}{\textit{Fixed-Length Retrieval Chunks}} \\
            \midrule
            Uniform Split (2 min)
            & 120.0 & \underline{40.2} & 65.7 & 75.0 & 40.4 & \underline{66.8} & \underline{75.6} & \textbf{52.3} & \underline{72.0} & 78.5 & 54.5 & 72.0 & 78.4 \\
            Uniform Split (3 min)
            & 180.0 & \textbf{43.8} & \underline{68.1} & \textbf{77.9} & \underline{42.6} & 65.2 & 75.6 & 51.1 & 69.4 & 76.3 & \underline{55.4} & 72.1 & 77.8 \\
            Uniform Split (4 min)
            & 240.0 & \textbf{43.8} & \textbf{68.5} & \underline{77.1} & \textbf{44.4} & \textbf{67.1} & \textbf{77.4} & 51.3 & 71.1 & 77.8 & 54.5 & 71.0 & 76.1 \\
            \midrule
            Oracle Uniform Clip (3 min)
            & 180.0
            & \multicolumn{3}{c}{64.8}
            & \multicolumn{3}{c}{70.4}
            & \multicolumn{3}{c}{51.8}
            & \multicolumn{3}{c}{57.4} \\
            \midrule
            \multicolumn{14}{l}{\textit{Scene-Based Retrieval Chunks}} \\
            \midrule
            PySceneDetect~\citep{castellano2024pyscenedetect}
            & 4.24 & 35.3 & 57.7 & 68.3 & 36.3 & 58.8 & 68.8 & 50.8 & 70.9 & 79.0 & 55.0 & \underline{73.0} & \underline{79.5} \\
            BaSSL~\citep{mun2022bassl}
            & 5.44 & 34.9 & 59.1 & 69.3 & 34.6 & 59.0 & 69.3 & \underline{51.5} & \textbf{72.5} & \textbf{79.3 }& \textbf{56.0} & \textbf{73.7} & \textbf{79.6} \\
            GenreDur~\citep{cho2026iclrsceneseg}
            & 54.1 & 37.5 & 62.0 & 72.2 & 36.9 & 61.3 & 70.9 & 51.1 & 71.1 & 77.9 & 54.8 & 71.8 & 77.1 \\
            Length-Controlled
            & 115.5 & 40.1 & 64.7 & 75.7 & 41.3 & 63.5 & 74.5 & 50.4 & 70.5 & 77.9 & 55.2 & 71.7 & 78.4 \\
            \bottomrule
        \end{tabular}%
    }
    \caption{Comparison of scene segmentation baselines on movie question answering results on MovieStory101 \citep{he2024storyteller}. We evaluate downstream QA models including Qwen3-4B, Qwen3-8B, InternVL3.5 8B, and InternVL3.5 14B while varying the retrieved scene units.}
    \label{tab:downstream_moviestory}
    \vspace{-8pt}
\end{table*}

\paragraph{Movie Question Answering.}
We next evaluate MovieStory101 \citep{he2024storyteller}, which measures movie question answering over three-minute story clips designed to preserve coherent local narrative context. Unlike MAD, this task depends less on pinpointing a brief moment and more on retrieving chunks that preserve character interactions and plot continuity. Following the same retrieval pipeline as in the previous experiments, we retrieve scene segments from the full-length movie using text queries and provide the retrieved segments as visual evidence to downstream QA models (Qwen~\cite{bai2025qwen3vl}, InternVL~\cite{wang2025internvl3_5}). Table~\ref{tab:downstream_moviestory} shows that uniform chunking remains a strong baseline, while existing scene-based segmentation methods do not show a consistent advantage. This is most evident for the Qwen models, where BaSSL and GenreDur underperform the best uniform baseline. For InternVL, scene-based chunks are often competitive, but still not clearly superior to fixed-length splitting. Together with the MAD results, this suggests that current scene segmentation methods are not reliably better retrieval units across downstream movie understanding tasks.

\paragraph{Movie Claim Verification.}
\begin{table*}[t]
    \centering
    \scriptsize
    \setlength{\tabcolsep}{4pt}
    \resizebox{0.8\textwidth}{!}{%
        \begin{tabular}{l c cccc cccc}
            \toprule
            \multirow{3}{*}{\textbf{Condition}}
            & \multirow{3}{*}{\textbf{Avg. Length (s)}}
            & \multicolumn{4}{c}{\textbf{Qwen3-VL (4B)}~\citep{bai2025qwen3vl}}
            & \multicolumn{4}{c}{\textbf{Qwen3-VL (8B)}~\citep{bai2025qwen3vl}} \\
            \cmidrule(lr){3-6} \cmidrule(lr){7-10}
            & & \multicolumn{4}{c}{\textbf{Pairwise Acc. (\%)}} 
            & \multicolumn{4}{c}{\textbf{Pairwise Acc. (\%)}} \\
            \cmidrule(lr){3-6} \cmidrule(lr){7-10}
            & & $K$=1 & $K$=2 & $K$=3 & $K$=5
            & $K$=1 & $K$=2 & $K$=3 & $K$=5 \\
            \midrule
            % \multicolumn{10}{l}{\textit{Whole-Movie Input}} \\
            % \midrule
            Full-length Movie
            & --
            & \multicolumn{4}{c}{12.56}
            & \multicolumn{4}{c}{23.77} \\
            \midrule
            \multicolumn{10}{l}{\textit{Fixed-Length Retrieval Chunks}} \\
            \midrule
            Uniform Split (2 min)
            & 120.0
            & \underline{12.90} & 20.51 & 25.23 & 33.29
            & 9.45 & 16.71 & 22.85 & 29.26 \\
            Uniform Split (3 min)
            & 180.0
            & 12.79 & 20.07 & \underline{25.35} & 32.14
            & 8.41 & \underline{16.82} & 22.70 & 28.00 \\
            Uniform Split (4 min)
            & 240.0
            & 12.56 & 19.93 & 24.19 & 30.65
            & 8.45 & 16.59 & 22.35 & 29.26 \\
            \midrule
            \multicolumn{10}{l}{\textit{Scene-Based Retrieval Chunks}} \\
            \midrule
            PySceneDetect~\citep{castellano2024pyscenedetect}
            & 15.8
            & 7.72 & 12.79 & 19.12 & 29.15
            & 6.91 & 12.90 & 17.97 & 26.27 \\
            BaSSL~\citep{mun2022bassl}
            & 10.0
            & 5.88 & 9.79 & 13.25 & 18.20
            & 4.95 & 9.68 & 12.90 & 17.05 \\
            GenreDur~\citep{cho2026iclrsceneseg}
            & 590.7
            & \underline{12.90} & \textbf{21.14} & 24.53 & \underline{36.98}
            & \underline{9.56} & \textbf{17.40} & \textbf{23.96} & \textbf{33.41} \\
            Length-Controlled
            & 107.5
            & \textbf{12.98} & \underline{21.00} & \textbf{27.72} & \textbf{41.47}
            & \textbf{12.21} & 16.43 & \underline{23.16} & \underline{31.62} \\
            \bottomrule
        \end{tabular}%
    }
    \caption{Impact of segmentation methods on movie fact verification in MF$^2$ \citep{zaranis2025mf2}. Whole-movie rows feed the entire video directly into the MLLM without retrieval. All other rows use Qwen3-VL-Embedding retrieval and pass the top-$K$ retrieved chunks to the downstream model. Uniform splitting remains a strong baseline, while comparisons between length-controlled and semantically grounded chunks reveal that narrative boundary alignment contributes beyond chunk length alone.}
    \label{tab:downstream_qa}
    \vspace{-6pt}
\end{table*}
Finally, we evaluate MF$^2$ \citep{zaranis2025mf2}, a long movie understanding benchmark built from full-length open-licensed films and fact-fib claim pairs. Following the protocol in~\citep{zaranis2025mf2}, we report pairwise accuracy, which measures whether a model prefers the true claim over its paired false claim given the same movie context. Table~\ref{tab:downstream_qa} shows that retrieval is preferable to feeding the whole movie directly, but scene-based chunking is not consistently superior to simple uniform splitting. This is the key point of the comparison. GenreDur remains competitive, and semantically motivated boundaries can help in some settings, but current scene segmentation methods do not show a clear or dominant advantage over fixed-length chunking. The gap between semantic validity and practical retrieval strength motivates the benchmark audit in the next section.

\paragraph{Takeaways.}
We highlight two findings from the downstream experiments. (i) Existing scene segmentation methods, including both trained models and classical computer-vision (PySceneDetect) baselines, are not dominant retrieval units for movie RAG; across tasks, they fail to show a consistent advantage over simple uniform splitting. (ii) On MovieStory101, the oracle 3-minute clips substantially outperform retrieved top-1 chunks. To better understand this gap, we manually examined 63 randomly sampled oracle clips out of 632 test clips (10\%) and found that all of them contained coherent narrative scenes, whereas segments predicted by existing scene segmentation methods often failed to maintain narrative coherence. This observation motivates the next section, where we more directly examine whether standard scene annotations align with narrative structure.

%% file: 03_our_dataset.tex
\vspace{-2mm}
\section{Why Existing Scene Segmentation Falls Short for Movie RAG}
\vspace{-2mm}
\label{sec:why}

Section~\ref{sec:modern_movie_rag} shows that current scene segmentation methods do not yield effective retrieval units for downstream movie understanding. Across temporal grounding, question answering, and fact verification, they often fail to outperform simple temporal fixed-length splitting. This suggests a mismatch between the boundaries emphasized by existing scene segmentation benchmarks and the narrative event units needed for movie RAG. We therefore first define the narrative scene boundaries required for movie RAG, and then revisit standard scene annotations to assess how well they align with this definition.

% A brief manual inspection of the core three-minute clips in MovieStory101 further motivates this view. In most cases, the provided clip preserves a coherent local narrative episode rather than a segment defined only by surface visual transition. We use this observation only as motivation. Our main evidence comes from an explicit scene boundary definition grounded in Event Segmentation Theory and a targeted audit of MovieNet-SSeg under that definition.

\subsection{Defining Scene Boundaries with Event Segmentation Theory (EST)}

\begin{figure}[t]
    \centering
    \includegraphics[width=\columnwidth]{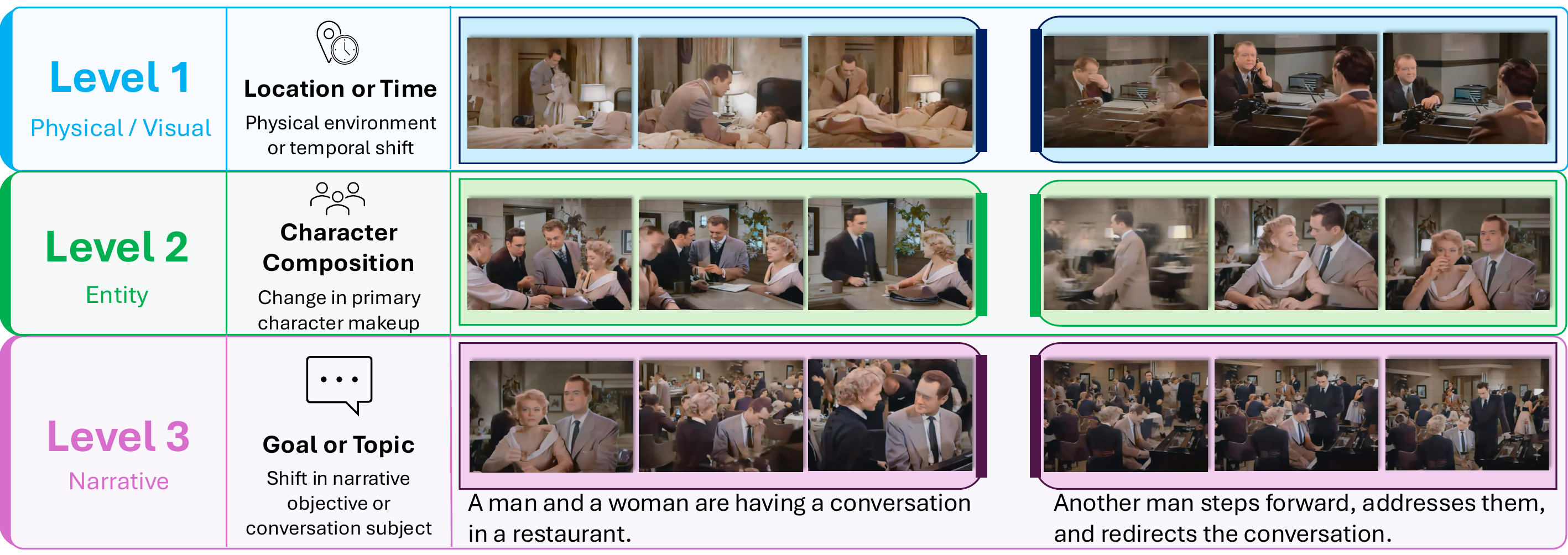}
    \vspace{-10pt}
    \caption{\textbf{Three levels of narrative-centric scene transitions.} We characterize scene segmentation candidate boundaries by changes in physical setting, character configuration, and narrative focus. A valid narrative scene boundary must include Level~3 change, while Level~1 and Level~2 cues alone are not sufficient.}
    \label{fig:levels}
    \vspace{-15pt}
\end{figure}

We define scene boundaries as meaningful narrative event changes rather than visually salient surface transitions. Existing scene segmentation annotations often include boundaries associated with cuts, shot transitions, or other local visual changes. We hypothesize that these boundaries are effective for capturing visually salient shifts, but less appropriate for evaluating whether a model can recognize when one coherent narrative event ends, and another begins.

Our definition is grounded in Event Segmentation Theory (EST), which argues that people parse continuous experience into discrete events by tracking changes in time, place, participants, causes, and goals \citep{zacks2007eventsegmentation,zacks2009segmentation}. This perspective is especially relevant to narrative film, where viewers do not segment experience at cuts alone and may maintain event continuity across visual changes when the underlying action remains coherent \citep{magliano2011eventsegfilm}. We therefore define scene boundaries in terms of event structure rather than perceptual discontinuity alone.

To operationalize this idea, we annotate three levels of scene transition, as illustrated in Figure~\ref{fig:levels}.

\vspace{-2mm}
\begin{itemize}[leftmargin=1.5em, itemsep=0.0em, label=\textbullet]
    \item \textbf{Level 1}: Physical changes in time, place, or other visually salient aspects of the scene.
    \item \textbf{Level 2}: Changes in the central character or entity configuration.
    \item \textbf{Level 3}: Shifts in the core goal, interaction, or conversational focus of the ongoing event.
\end{itemize}
\vspace{-2mm}

These levels are not mutually exclusive and may co-occur at the same timestamp. Our key design choice is that a valid narrative scene boundary must include a Level~3 transition. A candidate boundary may therefore satisfy Level~3 alone or together with Level~1 or Level~2, but any boundary without Level~3 is insufficient for defining a full scene transition. This criterion prevents the task from collapsing into visual chunking and focuses the analysis on narratively coherent event structure. Following EST, Level~1 and Level~2 track changes in time, place, and participants, while Level~3 captures the change in organizing goal or causal interaction that defines a coarse event \citep{kurby2008segmentation}. Requiring Level~3 is thus a task-specific operationalization of coarse retrievable events rather than a universal claim of EST.

\paragraph{Revisiting MovieNet-SSeg Annotations.}
\label{sec:revisit_movienet}
\begin{wraptable}{r}{0.35\columnwidth}
    \vspace{-13pt}
    \centering
    \small
    \setlength{\tabcolsep}{4pt}
    \begin{tabular}{lc}
        \toprule
        \textbf{Attribute} & \textbf{Value} \\
        \midrule
        % Movies audited & 10 \\
        % Boundaries audited & 1240 \\
        Level~1 & 80.4 \\
        Level~2 & 31.6 \\
        Level~3 & 11.4 \\
        None & 15.7 \\
        \bottomrule
    \end{tabular}
    \caption{\textbf{Audit of MovieNet-SSeg under our EST-based scene definition.} Values are percentages. Level~1, Level~2, and Level~3 are non-exclusive because a boundary may satisfy multiple levels. None denotes a benchmark boundary that satisfies none of the three criteria.}
    \label{tab:dataset_comparison}
    \vspace{-10pt}
\end{wraptable}
Most prior scene segmentation methods are trained and evaluated on MovieNet-SSeg \citep{huang2020movienet,mun2022bassl,wu2022scrl,cho2026iclrsceneseg}. To examine how its annotations relate to our EST-grounded definition, we revisit the MovieNet-SSeg test split and relabel a stratified sample of ten movies across genres using the three-level criteria above. Table~\ref{tab:dataset_comparison} summarizes the result. The audited MovieNet subset is dominated by Level~1 physical change, while only a small fraction of boundaries satisfy Level~3 narrative change. A nontrivial portion of benchmark boundaries also satisfies none of the three levels under our definition. This suggests that many MovieNet-SSeg boundaries align with visually salient transitions even when they do not correspond to full narrative scene transitions. Detailed results for each movie can be found in Appendix~\ref{app:movienet_audit_details}.

This does not make MovieNet-SSeg invalid for scene segmentation as originally defined. Rather, it indicates that the annotation target emphasized by existing scene segmentation practice differs from the one needed for movie RAG. As a result, models optimized for visually prominent boundaries may still produce retrieval chunks that fragment the narrative evidence required for downstream reasoning.

%% file: 04_baseline.tex
\vspace{-2mm}
\section{\ourdataset: Narrative-Centric Scene Segmentation Dataset}
\vspace{-2mm}
\label{sec:ourdataset}

\subsection{Dataset Construction: Building \ourdataset}
\label{sec:building_dataset}

\begin{wraptable}{r}{0.35\columnwidth}
\vspace{-2em}
\centering
\small
\setlength{\tabcolsep}{5pt}
\begin{tabular}{lrr}
\toprule
\textbf{Action} & \textbf{Count} & \textbf{Rate} \\
\midrule
Confirm & 1684 & 58.4\% \\
Relabel & 1027 & 35.6\% \\
Delete  & 99   & 3.4\% \\
Move    & 34   & 1.2\% \\
Insert  & 30   & 1.0\% \\
Hard    & 8    & 0.3\% \\
\bottomrule
\end{tabular}
\caption{\textbf{Cross-verification in \ourdataset.}}
\label{tab:verification_outcomes}
\vspace{-2.0em}
\end{wraptable}

Following the mismatch identified in Section~\ref{sec:modern_movie_rag}, we build \ourdataset to evaluate scene boundaries as narrative event changes in full-length movies rather than as visually salient breaks. We adopt the Event Segmentation Theory (EST)-grounded three-level scene boundary definition introduced in Figure~\ref{fig:levels}, where valid scene boundaries must include Level~3 narrative change \citep{zacks2007eventsegmentation,zacks2009segmentation}. This design targets the kind of coherent event units needed by movie RAG, where retrieval quality depends on preserving interactions, goals, and story progression rather than only local visual continuity.

The core of \ourdataset is built on the same movies in MF$^2$ \citep{zaranis2025mf2}, which provides full-length open-licensed movies together with an independent downstream movie reasoning task. We choose this source because it enables reproducible annotation and downstream evaluation on the same long-form narrative domain. In total, our annotation effort covers 53 full-length movies with an average duration of 88.3 minutes, yielding 2,371 narrative scenes.

Because semantic scene boundaries are inherently more ambiguous than physical cuts, we use a three-stage annotation protocol consisting of independent boundary proposal, structured cross-verification, and final consensus adjudication. During cross-verification, each candidate boundary receives one of six actions, namely confirm, relabel, delete, move, insert, or hard. Table~\ref{tab:verification_outcomes} summarizes the outcomes.

Most verification decisions either confirm an existing boundary or refine it through relabeling. Only a small minority require deletion, movement, or insertion, which suggests that disagreements usually concern boundary placement or interpretation rather than the existence of a narrative transition itself. After adjudication, every accepted boundary in \ourdataset retains a Level~3 narrative shift by construction. More details regarding the construction and statistics of \ourdataset are in Appendices~\ref{app:annotation_pipeline} and~\ref{app:additional_statistics}.

% \paragraph{\ourdataset statistics}
% \ourdataset is designed for long-horizon narrative understanding rather than clip-level scene parsing. The dataset spans full-length movies with diverse editing styles and yields a broad range of scene durations, from short transitions to substantially longer narrative units. Additional descriptive statistics, including duration histograms and label-combination breakdowns, are provided in Appendix~\ref{app:additional_statistics}.
\vspace{-5pt}
\subsection{Hand-Labeled Narrative Scenes Improve Movie RAG}
\vspace{-5pt}
\label{sec:oracle_movie_rag}

We next ask whether these hand-labeled narrative scenes improve downstream movie RAG when the boundary definition explicitly emphasizes narrative coherence. To isolate this question from automatic segmentation error, we compare fixed-length uniform chunks against semantic oracle boundaries on MF$^2$ claim verification \citep{zaranis2025mf2}. %Following MF$^2$, 
We report pairwise accuracy, which counts a prediction as correct when the model ranks the true claim above its paired false claim for the same movie.

\begin{table}[t]
    \centering
    \small
    \resizebox{0.8\textwidth}{!}{%
        \begin{tabular}{l cccc cccc}
            \toprule
            \multirow{2}{*}{\textbf{Condition}}
                & \multicolumn{4}{c}{\textbf{Qwen3-VL (4B)}}
                & \multicolumn{4}{c}{\textbf{Qwen3-VL (8B)}} \\
            \cmidrule(lr){2-5} \cmidrule(lr){6-9}
            & $K$=1 & $K$=2 & $K$=3 & $K$=5
            & $K$=1 & $K$=2 & $K$=3 & $K$=5 \\
            \midrule
            Uniform Split (2 min)
            & 12.90 & 20.51 & 25.23 & 33.29
            & 9.45 & 16.71 & 22.85 & 29.26 \\
            Uniform Split (3 min)
            & 12.79 & 20.07 & 25.35 & 32.14
            & 8.41 & 16.82 & 22.70 & 28.00 \\
            Uniform Split (4 min)
            & 12.56 & 19.93 & 24.19 & 30.65
            & 8.45 & 16.59 & 22.35 & 29.26 \\
            GenreDur \citep{cho2026iclrsceneseg}
            & 12.90 & 21.14 & 24.53 & 36.98
            & 9.56 & 17.40 & 23.96 & 33.41 \\
            \ourbaseline
            & \underline{13.13} & \textbf{22.47} & \underline{30.30} & \underline{42.86}
            & \underline{11.75} & \underline{19.59} & \underline{25.92} & \underline{35.37} \\
            \textbf{\ourdataset}
            & \textbf{18.66} & \underline{22.24} & \textbf{30.53} & \textbf{43.43}
            & \textbf{19.59} & \textbf{22.35} & \textbf{28.34} & \textbf{41.47} \\
            \bottomrule
        \end{tabular}%
    }
    \vspace{1mm}
    \caption{\textbf{\ourdataset improves pairwise accuracy on movie claim verification in MF$^2$.} All rows use the same RAG pipeline with a Qwen3-VL-Embedding retriever and Qwen3 QA models, differing only in the retrieval units used under each condition. We compare fixed-length uniform chunks against semantic oracle boundaries annotated under our EST-grounded definition.}
    \label{tab:oracle_vs_uniform}
    \vspace{-5mm}
\end{table}

\begin{table}[t]
    \centering
    \small
    \resizebox{1.0\textwidth}{!}{%
    \begin{tabular}{l ccccccc}
        \toprule
        \textbf{Segmentation} & \textbf{R1@0.1} & \textbf{R5@0.1} & \textbf{R1@0.3} & \textbf{R5@0.3} & \textbf{R1@0.5} & \textbf{R5@0.5} & \textbf{Avg.} \\
        \midrule
        Best Prior Scene Segmentation (Length-Controlled) & 8.19 & 21.75 & 3.38 & 11.29 & 2.25 & 6.49 & 8.89 \\
        \textbf{\ourdataset} & \textbf{12.95} & \textbf{27.59} & \textbf{8.44} & \textbf{18.24} & \textbf{5.30} & \textbf{11.20} & \textbf{13.79} \\
        \bottomrule
    \end{tabular}}
    \vspace{1mm}
    \caption{\textbf{\ourdataset generalizes to temporal grounding on 10 additional MAD movies.} We compare against the strongest baseline derived from prior scene segmentation methods on these movies, using the same retriever and TimeLens-8B localizer as Table~\ref{tab:downstream_mad} and differing only in the retrieval units.}
    \label{tab:mad10_oracle}
\end{table}

Table~\ref{tab:oracle_vs_uniform} shows that hand-labeled semantic boundaries consistently outperform all fixed-length baselines and the automatic GenreDur segmentation under the same retrieval pipeline. \ourdataset achieves the best pairwise accuracy in most model--retrieval settings, establishing that narratively valid chunk boundaries substantially improve downstream movie RAG when the boundaries are correct.
The gains over uniform splits and GenreDur are not marginal, as \ourdataset outperforms them in every setting, showing that retrieval quality depends not just on chunk size, but on whether boundaries align with coherent narrative units. These results provide direct evidence that retrieval quality in movie RAG is bottlenecked not only by the embedding model, but also by whether chunk boundaries preserve coherent narrative units.

To test whether the taxonomy is actionable without labels, we prompt Qwen3.5-9B zero-shot with our three-level definition (\ourbaseline, Appendix~\ref{app:zeroshot}), which raises R@0.9 on \ourdataset from 1.92 for the strongest existing method to 22.78 (Table~\ref{tab:scene_seg_era}) and yields the best non-oracle retrieval units in Table~\ref{tab:oracle_vs_uniform}. It still falls clearly short of the oracle at $K$=1, leaving \ourdataset far from saturated.

To test whether this benefit extends beyond MF$^2$ and claim verification, we annotate 10 additional MAD movies with the same protocol, selecting them before observing any results and without exposing annotators to MAD queries or ground-truth timestamps (Appendix~\ref{app:mad10}). As shown in Table~\ref{tab:mad10_oracle}, \ourdataset outperforms the strongest baseline derived from prior scene segmentation methods on every metric, raising the average from 8.89 to 13.79.

At the same time, these results should be interpreted as an oracle analysis rather than as evidence that current automatic scene segmentation methods already solve the problem. Instead, they establish a practical upper bound and motivate \ourdataset as a benchmark for testing whether future segmentation models can recover narrative boundaries that translate into measurable gains in downstream movie understanding.

%% file: 05_experiments.tex
\vspace{-10pt}
\section{Additional Discussion and Analyses}
\vspace{-6pt}
\label{sec:discussion}

\paragraph{Qualitative Analysis.}
\begin{figure}[t]
  \centering
  \includegraphics[width=\textwidth]{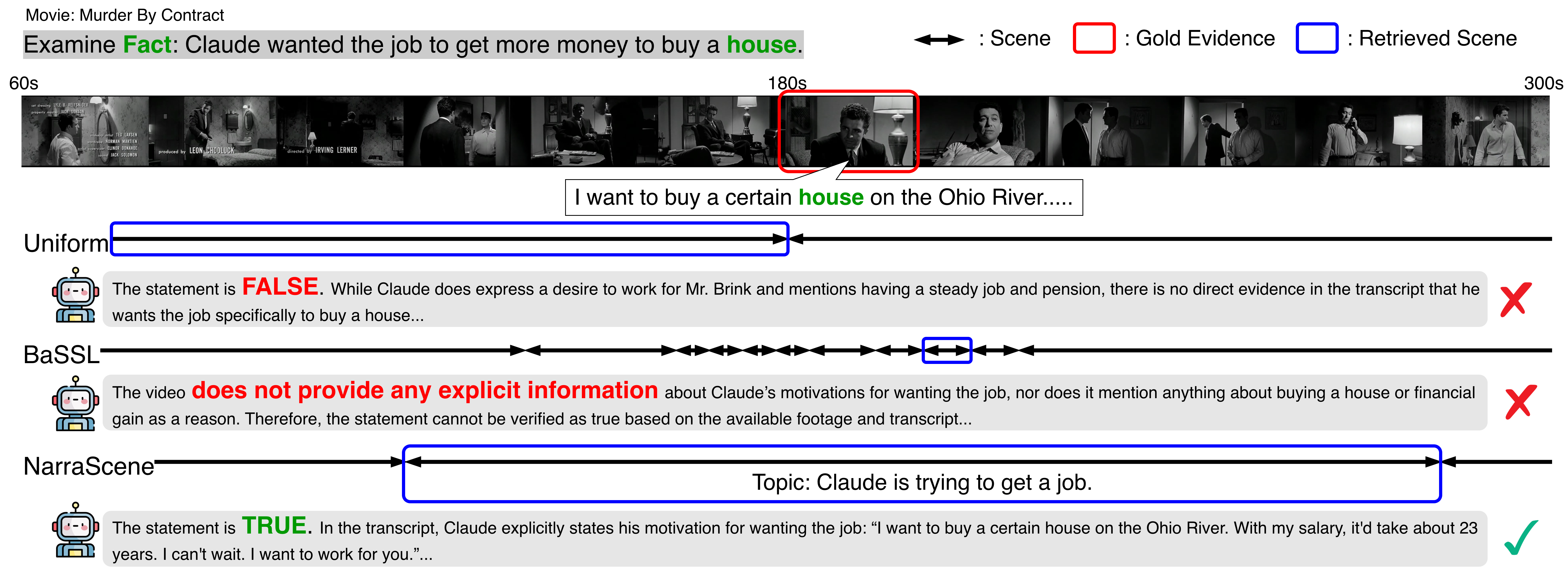}
  \vspace{-10pt}
  \caption{\textbf{Qualitative example of retrieved chunks under different scene segmentation strategies.}
  We compare fixed-length chunking, a standard scene segmentation method (BaSSL), and \ourdataset on a movie claim verification example from \textit{Murder by Contract}. The claim asks whether Claude wanted the job to get more money to buy a house. Uniform chunking and BaSSL retrieve incomplete scene segments and lead to an incorrect answer, while the narrative scene retrieved from \ourdataset preserves the full event and contains the decisive evidence, namely Claude's explicit statement that he wants to buy a certain house on the Ohio River.}
  \label{fig:qualitative_example}
  \vspace{-5pt}
\end{figure}
% \todo{add goal of scene in figure}

\begin{table}[t]
    \centering
    \resizebox{0.85\textwidth}{!}{%
        \begin{tabular}{lcccccc}
            \toprule
            \textbf{Method} & \textbf{mIoU} & \textbf{R@0.1} & \textbf{R@0.3} & \textbf{R@0.5} & \textbf{R@0.7} & \textbf{R@0.9} \\
            \midrule
            \multicolumn{7}{c}{\textit{Uniform Heuristics (Training-Free)}} \\
            \midrule
            Uniform Split (2 min)                & 49.3  & \textbf{98.8}  & \textbf{82.9}  & 48.5  & 15.3  & 1.7   \\
            Uniform Split (3 min)                & 44.5  & 97.1  & 74.0  & 38.8  & 12.5  & 1.8   \\
            Uniform Split (4 min)                & 38.8  & 94.1  & 61.8  & 28.0  & 10.3  & 1.0   \\
            \midrule
            \multicolumn{7}{c}{\textit{Prior Scene Segmentation Methods}} \\
            \midrule
            PySceneDetect~\citep{castellano2024pyscenedetect} & 34.92 & 90.40 & 51.04 & 22.47 & 9.06  & 1.92  \\
            BaSSL~\citep{mun2022bassl}           & 11.52 & 31.62 & 11.38 & 4.74  & 1.43  & 0.16  \\
            GenreDur~\citep{cho2026iclrsceneseg} & 12.43 & 29.69 & 11.10 & 6.33  & 3.56  & 1.42  \\
            \midrule
            \multicolumn{7}{c}{\textit{Narrative-Prompted (Training-Free)}} \\
            \midrule
            \ourbaseline                         & \textbf{54.14} & 92.91 & 71.15 & \textbf{51.29} & \textbf{34.63} & \textbf{22.78} \\
            \bottomrule
        \end{tabular}%
    }
    \vspace{2mm}
    \caption{\textbf{Existing scene segmentation methods fail to recover narrative boundaries.} All methods are evaluated with \ourdataset as ground truth. R@$t$ denotes the fraction of ground-truth scenes matched by a predicted scene with IoU $\geq t$.}
    \label{tab:scene_seg_era}
    \vspace{-10pt}
\end{table}

Figure~\ref{fig:qualitative_example} illustrates the main failure mode behind our quantitative results. The evidence needed to verify the claim is not distributed arbitrarily across time. It is concentrated in a narratively coherent exchange in which Claude explains why he wants the job. When segmentation fragments this exchange into shorter or visually defined chunks, the retriever can return only partial evidence, which makes the downstream verifier miss the key statement and answer incorrectly.

This example also clarifies why uniform chunking can remain a strong baseline while still failing to recover the most useful semantic unit. A fixed window may overlap part of the relevant exchange and therefore remain competitive in aggregate, but it does not explicitly preserve the event-level structure that makes the claim easy to verify. In contrast, the narrative scene groups the full conversational event into a single retrieval unit, so the model receives both the motivation and the supporting quote in one coherent context.

More broadly, this case highlights the transfer gap between legacy scene segmentation targets and downstream movie RAG. The decisive evidence in this example is not a visually salient transition, but a goal-relevant conversational reveal. This is precisely the type of boundary that is underemphasized when scene segmentation is treated mainly as detecting physical or local visual change. Our qualitative analysis therefore supports the central argument of the paper, namely that useful scene boundaries for movie RAG should be evaluated by the narrative evidence they preserve, not only by how well they match legacy boundary annotations.

\paragraph{Can Existing Methods Recover Narrative Boundaries?}
Table~\ref{tab:scene_seg_era} evaluates segmentation methods with \ourdataset as ground truth. Prior scene segmentation methods reach at most 34.92 mIoU and fall below even the 2-minute uniform split, showing that boundaries tuned to visual changes do not transfer to narrative event boundaries. At the strict R@0.9 threshold, all non-narrative methods stay below 2\%, indicating that none of them precisely recovers narrative scenes.

% \paragraph{How Well Does the Query-to-Scene Retriever Perform?}
\paragraph{How Retrieval Quality Affects Movie RAG?}

\begin{wraptable}{r}{0.46\columnwidth}
    \vspace{-20pt}
    \centering
    \small
    \setlength{\tabcolsep}{4pt}
    \caption{\textbf{Retrieval quality analysis on MovieStory101.} We compare actual top-1 retrieval against the chunk with the largest overlap with the annotated 3-minute evidence clip under Qwen3-VL 8B.}
    \label{tab:retrieval_quality}
    \begin{tabular}{lcc}
        \toprule
        \textbf{Setting} & \textbf{Top-1} & \textbf{Best Overlap} \\
        \midrule
        Uniform 3 min   & 41.6 & 60.9 \\
        GenreDur        & 36.2 & 50.4 \\
        \bottomrule
    \end{tabular}
    \vspace{-5pt}
\end{wraptable}

% Our RAG pipeline depends on Query-to-Scene retrieval to select useful chunks from a segmented movie. 
While the primary objective of this study is to analyze segmentation quality for movie RAG, we additionally investigate retriever quality, as retrieval performance is a key contributing factor to the overall results. We conduct this study on MovieStory101 as it provides gold evidence annotations of scene segmentation for each query. We replace the actual top-1 retrieved chunk with the chunk that has the largest overlap with the annotated 3-minute gold evidence clip, using Qwen3-VL 8B for both uniform chunking and GenreDur.

Table~\ref{tab:retrieval_quality} shows a large gap between actual top-1 retrieval and the best-overlap chunk. Replacing the top-1 chunk with the best-overlap chunk improves QA accuracy from 41.6 to 60.9 for uniform chunking and from 36.2 to 50.4 for GenreDur, indicating that end-to-end performance is limited by retrieval errors as well as chunk construction. The higher best-overlap accuracy of uniform chunking further suggests that chunk quality remains important even under favorable retrieval.

These results clarify the main findings. Lower downstream QA accuracy can arise because the retriever fails to select the best chunk, because the chunk itself does not preserve the right evidence, or both. Scene segmentation for movie RAG should therefore be evaluated through both retrieval quality and chunk quality rather than final QA accuracy alone.

% We are completing this analysis across all segmentation settings and will include full results in the final version. Together with the oracle analysis above, these results suggest that evaluating scene segmentation as a retrieval unit requires tracking both retrieval accuracy and chunk construction quality rather than relying on end-to-end task accuracy alone.

%% file: 06_conclusion.tex
\vspace{-2mm}
\section{Conclusion}
\vspace{-2mm}

In this work, we revisited scene segmentation as a way to define retrieval units for movie RAG. We observe that existing scene segmentation methods, despite their aim to preserve narrative coherence, do not consistently outperform simple uniform splitting on downstream movie understanding tasks. Our further analysis indicates that this gap originates from existing segmentation practice, where scene boundary annotations are more closely aligned with visually salient physical transitions than with narrative event change, making them ill-suited for movie RAG. Motivated by this mismatch, we built \ourdataset, a narrative-centric scene segmentation dataset for full-length movies grounded in Event Segmentation Theory. \ourdataset defines scene boundaries through physical, character, and narrative change, while requiring every valid boundary to include a narrative-level shift. Finally, we show that narrative-centric scene boundaries can outperform uniform chunking on MF$^2$ claim verification, suggesting that these boundaries are useful for movie RAG in principle. Taken together, our results suggest that scene segmentation for movie RAG should be evaluated not only by boundary detection accuracy under legacy benchmarks, but by its ability to recover narrative evidence units that improve downstream movie understanding.

%% file: 99_appendix.tex
\startcontents[appendices]

% 1. 그림과 표의 카운터가 섹션에 종속되지 않고 전체에서 이어지도록 설정
\counterwithout{figure}{section}
\counterwithout{table}{section}

% 2. 그림과 표의 번호를 0으로 초기화
\setcounter{figure}{0}
\setcounter{table}{0}

% 3. 그림과 표의 번호 출력 형식을 알파벳(A, B, C...)으로 변경
\renewcommand{\thefigure}{\Alph{figure}}
\renewcommand{\thetable}{\Alph{table}}

\startcontents[appendices]
\section*{Appendix Contents}
\printcontents[appendices]{section}{1}{}

\section{Related Works}

\paragraph{Movie scene segmentation.}
Prior work on movie scene segmentation has focused on improving boundary prediction on established benchmarks such as MovieNet~\citep{huang2020movienet} and its recent extensions, using self-supervised, transformer-based, multimodal, and vision-language approaches~\citep{mun2022bassl,chen2021shotcol,bertasius21timesformer,islam2023trans4mer,sadoughi2023mega,berman2025scenevlm,ventura2025chapterllama}. However, this literature largely treats the benchmark target itself as fixed. Our work revisits that assumption. Instead of proposing another boundary detector, we ask whether existing scene segmentation targets define the right context unit for downstream movie understanding. We show that the dominant target is more strongly aligned with visually salient transitions than with narrative event change, and introduce a narrative-centric dataset designed for downstream movie RAG.

\paragraph{Event segmentation in cognitive science.}
% \dk{this section may be merged with the above section.}
Event segmentation has been studied in cognitive science for decades~\citep{zacks2007eventsegmentation,kurby2008segmentation}. People parse continuous experience into discrete events by tracking changes in goals, causes, characters, and locations, and event boundaries arise when ongoing predictions fail~\citep{zacks2007eventsegmentation,zacks2009segmentation}. This process is hierarchical. Coarse event boundaries are associated with changes in goals and higher-level situation structure, whereas finer boundaries often reflect lower-level physical or action changes~\citep{zacks2001perceiving,kurby2008segmentation}. In narrative film, viewers do not segment experience at visual cuts alone. They also segment when the underlying situation or goal structure changes, even when perceptual discontinuities are weak or absent~\citep{magliano2011eventsegfilm,zacks2010brain}. This perspective is largely missing from existing movie scene segmentation benchmarks~\citep{huang2020movienet,cho2026iclrsceneseg} and from recent MLLM-based long-video systems that use scene-like units for retrieval and reasoning~\citep{zeng2025scenerag}. Our work uses Event Segmentation Theory to make this distinction explicit, to audit prior scene annotations, and to define \ourdataset around narrative-valid scene boundaries whose downstream utility can be tested directly in movie RAG.

\paragraph{RAG for long video understanding.}
Recent long-video understanding methods have increasingly moved away from feeding entire videos into a single MLLM context and instead rely on either context compression or retrieval-based evidence selection \citep{shu2025videoxl}. A growing line of work augments video LLMs with retrieval at the level of frames, clips, documents, or semantically segmented scenes, showing that explicit evidence selection is often more effective than brute-force context scaling for long-form reasoning \citep{tan2025ragadapter, jeong2025videorag,ma2025drvideo, zeng2025scenerag, luo2025videoragvis}. Our work is closely related to this trend, but focuses on a more fundamental question. Instead of proposing a new retrieval architecture, we study what constitutes an appropriate retrieval unit for movie understanding, and argue that cognitively grounded narrative scene boundaries provide a stronger substrate for downstream retrieval, localization, and reasoning than fixed temporal chunks or visually defined segments.

% \paragraph{Movie Understanding (Benchmarks).} \dk{movie related downstream tasks. But we may omit this as it is explained in the following section.}

\section{Detailed Downstream Task Experimental Setup}
\label{app:downstream_task_details}

\paragraph{Common experimental protocol.}
We keep four design choices fixed across all downstream tasks. First, the same scene candidates produced by each segmentation method are reused as the retrieval pool. Second, retrieval uses a single shared backbone, Qwen3-VL-Embedding-2B, with fixed clip preprocessing at 1\,FPS, $\ell_2$-normalized embeddings, and dot-product similarity. Third, within each task, all compared answer, localization, or verification models use identical retrieval outputs, so any downstream performance gap reflects only the non-retrieval model. Fourth, we follow the reference hyperparameter settings of the original works whenever available, and we do not tune them separately for each segmentation method. We evaluate on the official test split of MovieStory101 and MAD. We evaluate on the public benchmark of MF$^2$, which does not define a train and test split. All models use greedy decoding and fixed retrieval outputs. Reported results therefore come from a single deterministic run, except for minor infrastructure nondeterminism from GPU execution and vLLM scheduling.

\subsection{MovieStory101: Movie Question Answering}
\label{app:downstream_moviestory}

We evaluate Qwen3-VL-4B-Instruct, Qwen3-VL-8B-Instruct, InternVL3.5-8B, and InternVL3.5-14B as the QA models. All are served with vLLM under matched decoding settings with \texttt{temperature} = 0.0, \texttt{max\_tokens} = 1024, and Qwen's chain-of-thought channel disabled. All checkpoints use the same retrieval outputs, so differences reflect only the answering model.

Each MovieStory101 instance is a five-way multiple-choice question $(\text{A}, \text{B}, \text{C}, \text{D}, \text{E})$, where option E is a fixed ``I don't know'' fallback. We use a two-stage retrieve-then-answer protocol. In the retrieval stage, every scene clip produced by the segmentation method in Table~\ref{tab:downstream_moviestory} is encoded with Qwen3-VL-Embedding-2B under vLLM pooling mode with \texttt{dtype} = \texttt{bfloat16}. Clips are uniformly sampled at 1\,FPS with up to 64 frames each. The question is encoded as a single text query using the instruction \textit{``Retrieve the most relevant video clip that answers the given question.''} Candidate scenes are ranked by the dot product between $\ell_2$-normalized text and video embeddings, and the top 5 scenes are kept as evidence. In the answer stage, each retrieved scene is shown to the QA model independently together with the prompt in Figure~\ref{fig:moviestory_qa_prompt}. We parse the bracketed \textbf{[Answer]} tag from the model output. Responses without a parseable option letter are counted as incorrect.

Following the standard MovieStory101 protocol, we report \emph{Top-K accuracy} for $K \in \{1, 3, 5\}$. This metric is the fraction of questions for which at least one of the top-$K$ retrieved scenes yields the correct option. Each QA model is evaluated once on the same retrieval JSON.

\begin{figure}[h]
    \centering
    \begin{tcolorbox}[title={Multiple-Choice QA Prompt for MovieStory101}, colback=gray!2, colframe=black!30, arc=1mm, boxrule=0.4pt, width=0.95\textwidth, fonttitle=\bfseries, fontupper=\small]
    \textbf{[Multiple-Choice Question]}\\
    \texttt{\{question\}}

    Please answer the \textbf{[Multiple-Choice Question]} by watching the video. Only one of the options (A, B, C, D, E) is correct. Your response should follow this format.

    \textbf{[Reason]} Explain your reasoning.\\
    \textbf{[Answer]} Generate only one character from the options (A, B, C, D, E).
    \end{tcolorbox}
    \caption{Prompt template used to elicit a single-choice answer from the QA model on MovieStory101. The placeholder \texttt{\{question\}} is filled per instance.}
    \label{fig:moviestory_qa_prompt}
\end{figure}

\subsection{MAD: Movie temporal grounding}
\label{app:downstream_mad}

We evaluate TimeLens-8B and Vidi-7B as the two localizers. Both use greedy decoding with a per-model \texttt{max\_new\_tokens} taken from each model's reference setup. TimeLens-8B ingests scenes at 1\,FPS and uses beam search when more than one prediction per query is requested. Vidi-7B takes paired video and audio inputs through a SigLIP-so400m vision tower and a companion audio encoder, reflecting its multimodal design.

Each MAD instance is a free-form caption to be temporally grounded inside its movie. We use a two-stage retrieve-then-localize protocol. In the retrieval stage, every scene clip produced by the segmentation method in Table~\ref{tab:downstream_mad} is encoded with Qwen3-VL-Embedding-2B at 1\,FPS with up to 64 frames per clip and \texttt{max\_length} = 8192. The caption is encoded as a single text query. Scenes are ranked by the dot product between $\ell_2$-normalized text and video embeddings, and the top 5 scenes are kept as candidates. In the localization stage, each candidate is fed to the localizer independently. TimeLens-8B is queried with the prompt in Figure~\ref{fig:mad_localisation_prompts} and predicts boundaries directly in absolute seconds. Vidi-7B is queried with the corresponding prompt in the same figure and predicts a normalized \texttt{start\%--end\%} range, which we remap to seconds using the clip duration. Unparseable outputs are assigned an invalid temporal span and therefore receive zero IoU. Predictions are then mapped from the local clip timeline back to the full MAD movie timeline before scoring.

\begin{figure}[h]
    \centering
    \begin{tcolorbox}[title={Localization Prompts for MAD}, colback=gray!2, colframe=black!30, arc=1mm, boxrule=0.4pt, width=0.95\textwidth, fonttitle=\bfseries, fontupper=\small]
    \textbf{TimeLens-8B (absolute-second output).}\\
    Please find the visual event described by the sentence `\texttt{\{query\}}', determining its starting and ending times. The format should be `The event happens in \texttt{<start time>}~--~\texttt{<end time>} seconds'.

    \medskip

    \textbf{Vidi-7B (normalized-percentage output).}\\
    Given the frames from a video, answer the time range in percentage that corresponds to the query text split by a comma. Video length is \texttt{\{video\_length\}} and text query is \texttt{\{query\}}.
    \end{tcolorbox}
    \caption{Prompt templates used by the two MAD localizers. One prompt is issued per retrieved scene. The placeholders \texttt{\{query\}} and \texttt{\{video\_length\}} are filled per query with the MAD caption and the per-clip duration in seconds, respectively. TimeLens-8B emits boundaries directly in seconds. Vidi-7B emits a normalized \texttt{start\%--end\%} range, which we remap to seconds using the clip duration before mapping to the MAD movie timeline.}
    \label{fig:mad_localisation_prompts}
\end{figure}

\subsection{MF\texorpdfstring{$^2$}{2}: Movie Claim Verification}
\label{app:downstream_mf2}

We evaluate Qwen3-VL-4B-Instruct and Qwen3-VL-8B-Instruct as the two verifier models. Both are served with vLLM under matched decoding settings with greedy decoding and \texttt{max\_tokens} = 512. Videos are decoded at 1\,FPS with a maximum of 180 frames per scene. For the BaSSL-original treatment, where individual scenes are very short, the cap is reduced to 16 frames to avoid trivial up-sampling of repeated frames. Both checkpoints use the same retrieval outputs, so differences reflect only the verifier model.

Each MF$^2$ instance is a paired $(\texttt{true\_claim}, \texttt{false\_claim})$ statement about a single movie. We use a two-stage retrieve-then-verify protocol. In the retrieval stage, every scene clip produced by the segmentation method is encoded with Qwen3-VL-Embedding-2B at 1\,FPS with up to 180 frames per clip. To better simulate a practical scenario, we do not combine the paired statements into a single artificial query. Instead, we encode the \texttt{true\_claim} and the \texttt{false\_claim} as independent text queries and compute their respective similarity scores against each candidate scene. The final retrieval score for a given scene is then defined as the maximum of these two similarity values. To exploit the per-clip transcripts produced by Qwen3-ASR-1.7B, Qwen3-ForcedAligner-0.6B~\citep{shi2026qwen3asr}, and pyannote diarization~\citep{Plaquet23, Bredin23}, the ASR text of each candidate clip is appended to its visual input so that scenes are ranked by joint vision-language similarity. The ASR pipeline is identical for every segmentation method, so transcript availability does not confound the comparison. For each claim pair, we retain the top-$K$ retrieved scenes with $K \in \{1, 2, 3, 5\}$ as evidence.

In the verification stage, each claim is evaluated independently without reference to its partner statement or the pair label. We use a frozen MLLM to judge each claim using an OR ensemble over the top-$K$ retrieved scenes. A claim is judged \texttt{TRUE} if at least one retrieved scene yields a \texttt{TRUE} verdict, and \texttt{FALSE} if at least one yields a \texttt{FALSE} verdict. Each scene clip is normalized to 1\,FPS with a 180-frame ceiling. We parse the final verdict from the bracketed \texttt{<answer>} tag using the prompt in Figure~\ref{fig:mf2_verification_prompt}. In cases where a video fails to decode, the model falls back to a transcript-only prompt that omits the video token while maintaining the same answer format.

To assess the model's ability to distinguish fact from fiction, we report \emph{pairwise accuracy} at $K \in \{1, 2, 3, 5\}$ following the official MF$^2$ protocol. A pair is counted as correct only if at least one of the top-$K$ retrieved scenes yields a pairwise-consistent verdict. This requires that the same specific scene context correctly identifies the true statement as \texttt{TRUE} and the false statement as \texttt{FALSE}. We do not combine complementary verdicts across different retrieved scenes, as the goal is to evaluate whether the model can recover a coherent narrative event that supports the correct verdict. Results are reported separately for each $K$ and for both the 4B and 8B models.

\begin{figure}[h]
    \centering
    \begin{tcolorbox}[title={Claim Verification Prompt for MF$^2$}, colback=gray!2, colframe=black!30, arc=1mm, boxrule=0.4pt, width=0.95\textwidth, fonttitle=\bfseries, fontupper=\small]
    You are provided with a movie scene clip and its corresponding transcript. Your task is to carefully watch the video, read the transcript, and then determine whether the statement is true or false.\\
    Answer \textbf{TRUE} if the statement is true in its entirety based on the video and the transcript.\\
    Answer \textbf{FALSE} if any part of the statement is false based on the video and the transcript.

    \textbf{Transcript} \texttt{\{transcripts\}}

    \textbf{Statement} \texttt{\{claim\}}

    Based on the video and the transcript, is the above statement \textbf{TRUE} or \textbf{FALSE}?\\
    First provide an explanation of your decision-making process in at most one paragraph, and then provide your final answer in the format\\
    \texttt{<answer>TRUE</answer>} or \texttt{<answer>FALSE</answer>}.
    \end{tcolorbox}
    \caption{Prompt template used by the MF$^2$ verifier. One instance is run independently per retrieved scene and combined with an OR ensemble across the top-$K$ candidates. The placeholders \texttt{\{transcripts\}} and \texttt{\{claim\}} are filled with the ASR transcript of the scene and the claim under evaluation, respectively.}
    \label{fig:mf2_verification_prompt}
\end{figure}

\section{Film-wise MovieNet Audit Details}
\label{app:movienet_audit_details}

\begin{table*}[t]
  \centering
  \small
  \setlength{\tabcolsep}{5pt}
  \resizebox{\textwidth}{!}{%
    \begin{tabular}{lccccc}
      \toprule
      Film & \# Boundaries & L1 & L2 & L3 & None \\
      \midrule
      The Terminator (1984) & 48  & 89.6 (43)  & 54.2 (26)  & 10.4 (5)  & 10.4 (5)  \\
      Sherlock Holmes (2009) & 214 & 78.5 (168) & 33.2 (71) & 9.3 (20)  & 17.3 (37) \\
      Ring (1998) & 81 & 84.0 (68) & 33.3 (27) & 12.3 (10) & 11.1 (9) \\
      Gravity (2013) & 46 & 80.4 (37) & 4.3 (2) & 19.6 (9) & 13.0 (6) \\
      The Curious Case of Benjamin Button (2008) & 185 & 72.4 (134) & 25.9 (48) & 13.0 (24) & 26.5 (49) \\
      Insidious (2010) & 121 & 90.1 (109) & 33.9 (41) & 10.7 (13) & 9.1 (11) \\
      Men in Black 3 (2012) & 74 & 79.7 (59) & 32.4 (24) & 14.9 (11) & 9.5 (7) \\
      Source Code (2011) & 102 & 78.4 (80) & 25.5 (26) & 14.7 (15) & 17.6 (18) \\
      The Truman Show (1998) & 106 & 79.2 (84) & 35.8 (38) & 12.3 (13) & 19.8 (21) \\
      Inception (2010) & 263 & 81.7 (215) & 33.8 (89) & 8.0 (21) & 12.2 (32) \\
      \midrule
      \textbf{10 Movies Total} & \textbf{1240} & \textbf{80.4 (997)} & \textbf{31.6 (392)} & \textbf{11.4 (141)} & \textbf{15.7 (195)} \\
      \bottomrule
    \end{tabular}%
  }
  \caption{Film-wise audit of MovieNet scene boundaries using our cognitive taxonomy. Values are percentages, with raw boundary counts in parentheses.}
  \label{tab:movienet_audit}
\end{table*}

This appendix reports the film-wise results of the MovieNet audit summarized in Section~\ref{sec:revisit_movienet}. We re-annotated a stratified sample of ten films from the MovieNet-318 test set using our cognitive taxonomy and report the percentage and raw count of boundaries associated with each label type. The aggregate statistics in the main text are not driven by a single outlier title; rather, the film-wise results consistently show a strong concentration of Level~1 boundaries, relatively low Level~3 coverage, and a non-negligible proportion of \textit{None} cases across the sampled titles.

\section{Detailed \ourdataset Annotation Pipeline}
\label{app:annotation_pipeline}

\begin{figure*}[t]
    \centering
    \begin{subfigure}[t]{0.31\textwidth}
        \centering
        \includegraphics[width=\textwidth]{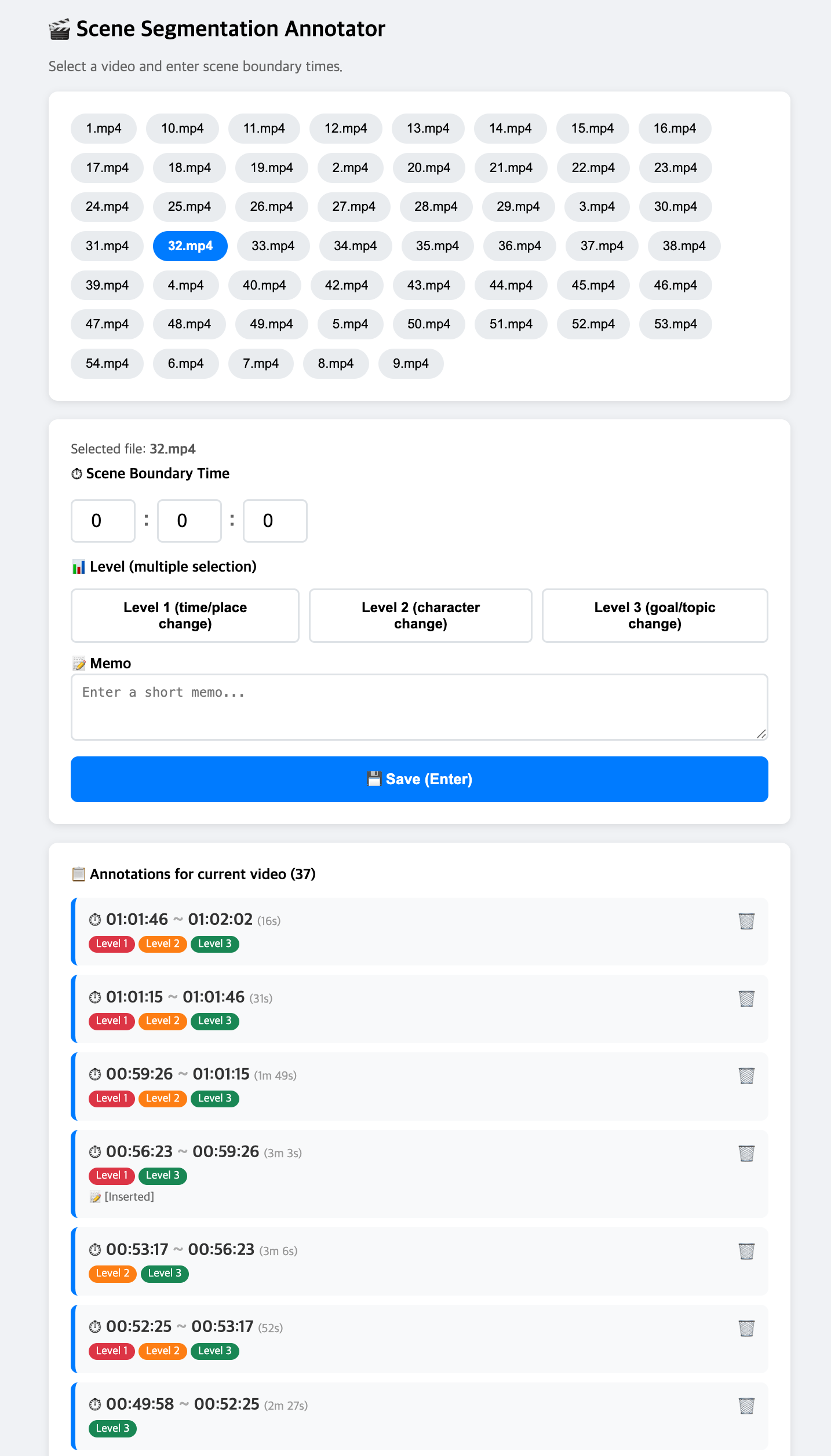}
        \caption{Stage 1: initial proposal interface.}
        \label{fig:annotation_stage1}
    \end{subfigure}
    \hfill
    \begin{subfigure}[t]{0.31\textwidth}
        \centering
        \includegraphics[width=\textwidth]{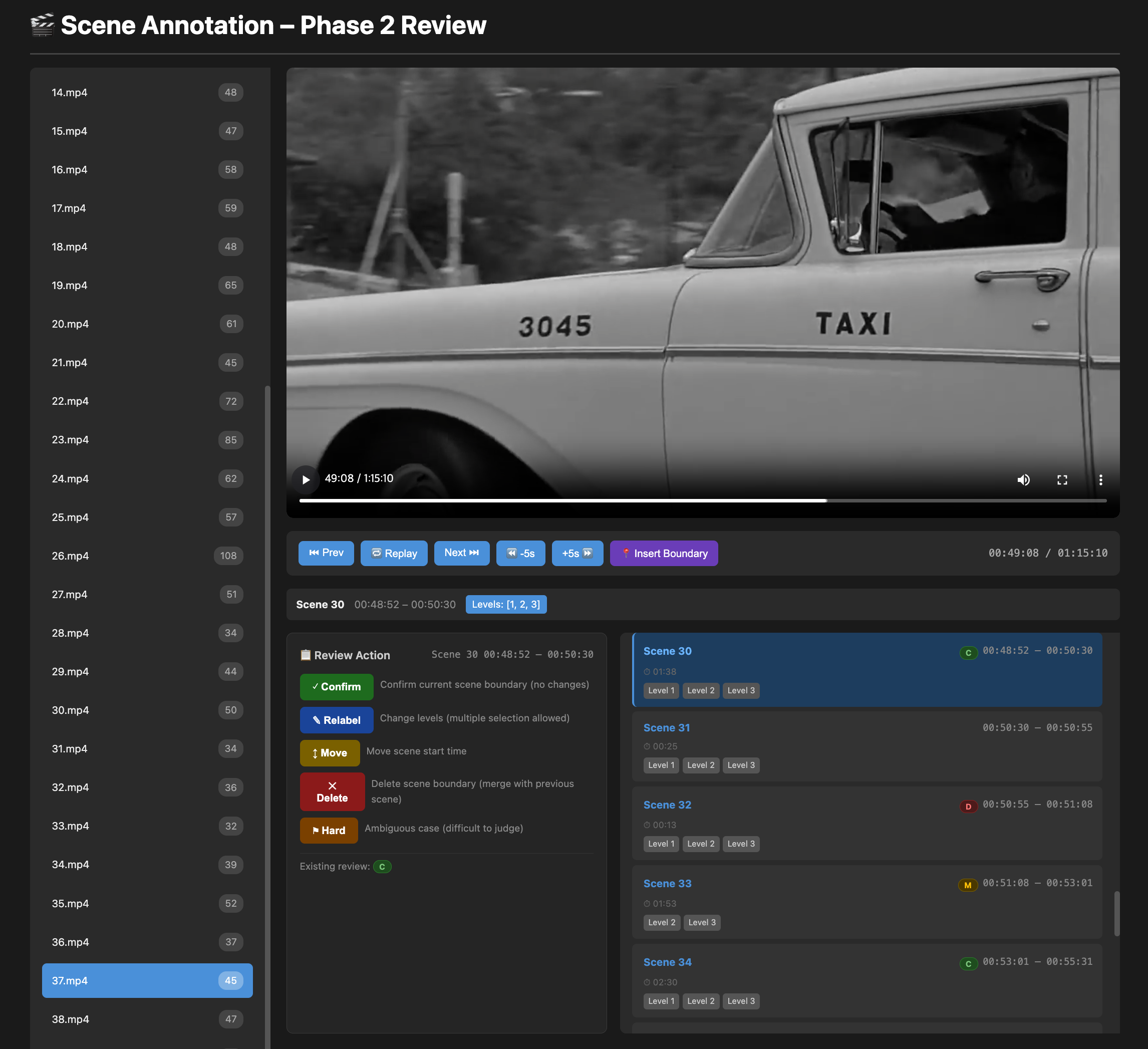}
        \caption{Stage 2: cross-verification interface.}
        \label{fig:annotation_stage2}
    \end{subfigure}
    \hfill
    \begin{subfigure}[t]{0.31\textwidth}
        \centering
        \includegraphics[width=\textwidth]{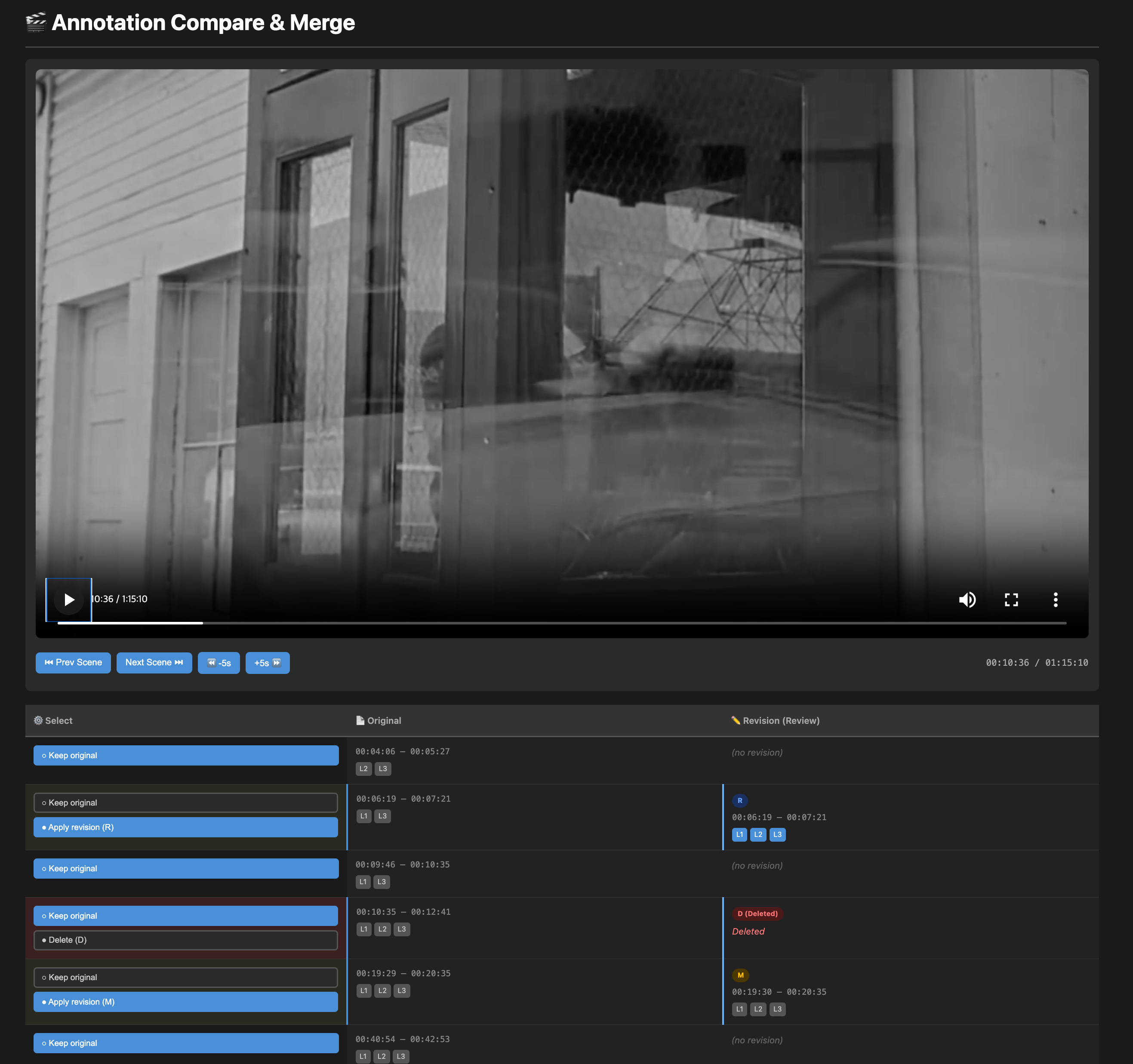}
        \caption{Stage 3: compare-and-merge consensus interface.}
        \label{fig:annotation_stage3}
    \end{subfigure}
    \caption{\textbf{Custom annotation interfaces used in the three-stage \ourdataset pipeline.} \textbf{(a)} In Stage 1, a single annotator watches the full movie, records scene boundary timestamps, assigns one or more narrative levels, and leaves a memo when needed. \textbf{(b)} In Stage 2, a second annotator reviews local clips and applies actions such as confirm, relabel, move, delete, insert, or hard marking for ambiguous cases that should be revisited in consensus. \textbf{(c)} In Stage 3, all four annotators compare the original and revised annotations and resolve disagreements through consensus. Zoom in for details.}
    \label{fig:annotation_pipeline}
\end{figure*}

\ourdataset was annotated by four annotators using a custom web-based annotation interface. For each movie, annotation began with a single-annotator proposal pass, where one annotator watched the full movie continuously and proposed narrative scene boundaries under our EST-grounded definition. As shown in Figure~\ref{fig:annotation_stage1}, the stage-1 interface allowed the annotator to select a movie, enter a boundary timestamp, assign one or more narrative levels, and leave a short memo for difficult cases. We adopted full-movie viewing rather than clip-first annotation so that annotators could preserve global story context, including causal continuity, character goals, and long-range callbacks, while using not only the visual stream but also audio and subtitles to decide whether a boundary marked a true narrative scene transition.

Each movie was then cross-checked by a second annotator in a dedicated review stage. In this phase, the reviewer inspected shorter clips around the proposed boundaries and decided whether each boundary should be confirmed (kept unchanged), relabeled (its Level 1, 2, and/or 3 annotation revised), deleted (the boundary removed), moved (the boundary timestamp adjusted), or supplemented by an inserted boundary (a new boundary added). Particularly ambiguous cases were marked as hard, indicating that the reviewer considered the case difficult enough to warrant later group discussion in the consensus stage. Figure~\ref{fig:annotation_stage2} shows that the stage-2 review interface exposed these action types explicitly and supported local replay around the boundary, allowing the reviewer to refine timestamp precision without discarding the global narrative interpretation established in the proposal stage. The resulting annotation records store explicit scene indices, start and end timestamps, and level labels, while the review logs preserve the revision actions for later auditing.

All disagreements from cross-verification were resolved in a final four-way consensus stage involving all annotators. In this stage, the team jointly re-watched each disputed segment, compared their narrative rationales, and produced a single agreed-upon decision for both the boundary location and its level label. As illustrated in Figure~\ref{fig:annotation_stage3}, the stage-3 compare-and-merge interface presented the original annotation alongside the review revision so that the annotators could directly reconcile conflicting decisions before finalizing the output. This final consensus process was especially important for hard cases such as gradual transitions, montage-like sequences, interleaved actions, or boundaries whose interpretation depended more on dialogue, subtitle content, or audio cues than on visible scene changes alone. For such cases, annotators decided whether the material remains subordinate to the current organizing objective or begins a self-contained narrative event. Montages and cross-cut sequences were kept as one scene when they form one causally integrated event, explanatory flashbacks were kept within the current scene unless they begin a self-contained episode, and sound bridges were treated as supporting but never sufficient evidence.

\section{Additional \ourdataset Statistics}
\label{app:additional_statistics}

\begin{table}[t]
    \centering
    \small
    \resizebox{\columnwidth}{!}{%
    \begin{tabular}{rrrrrrrrr}
        \toprule
        \# Movies & Hours & Mean movie & Median movie & \# Boundaries & \# Scenes & Mean scenes & Mean scene & Median scene \\
                  &       & dur. (min) & dur. (min)   &               &          & per movie   & dur. (sec) & dur. (sec) \\
        \midrule
        53 & 78.02 & 88.33 & 87.06 & 2371 & 2371 & 44.74 & 118.39 & 99.00 \\
        \bottomrule
    \end{tabular}%
    }
    \vspace{6pt}
    \caption{Detailed statistics of the MF$^2$-aligned portion of \ourdataset.}
    \label{tab:appendix_dataset_stats}
\end{table}

\begin{table}[t]
    \centering
    \small
    \setlength{\tabcolsep}{7pt}
    \begin{tabular}{lrr}
        \toprule
        Label combination & Count & Rate \\
        \midrule
        L1+L2+L3 & 1266 & 53.40\% \\
        L1+L3    & 489  & 20.62\% \\
        L2+L3    & 403  & 17.00\% \\
        L3       & 213  &  8.98\% \\
        \midrule
        L1 presence rate & \multicolumn{2}{r}{74.02\%} \\
        L2 presence rate & \multicolumn{2}{r}{70.39\%} \\
        L3 presence rate & \multicolumn{2}{r}{100.0\%} \\
        \bottomrule
    \end{tabular}
    \vspace{6pt}
    \caption{Detailed statistics of scene-boundary labels in the MF$^2$-aligned portion of \ourdataset.}
    \label{tab:appendix_boundary_stats}
\end{table}

\begin{figure}[t]
    \centering
    \begin{minipage}[t]{0.48\linewidth}
        \centering
        \includegraphics[width=\linewidth]{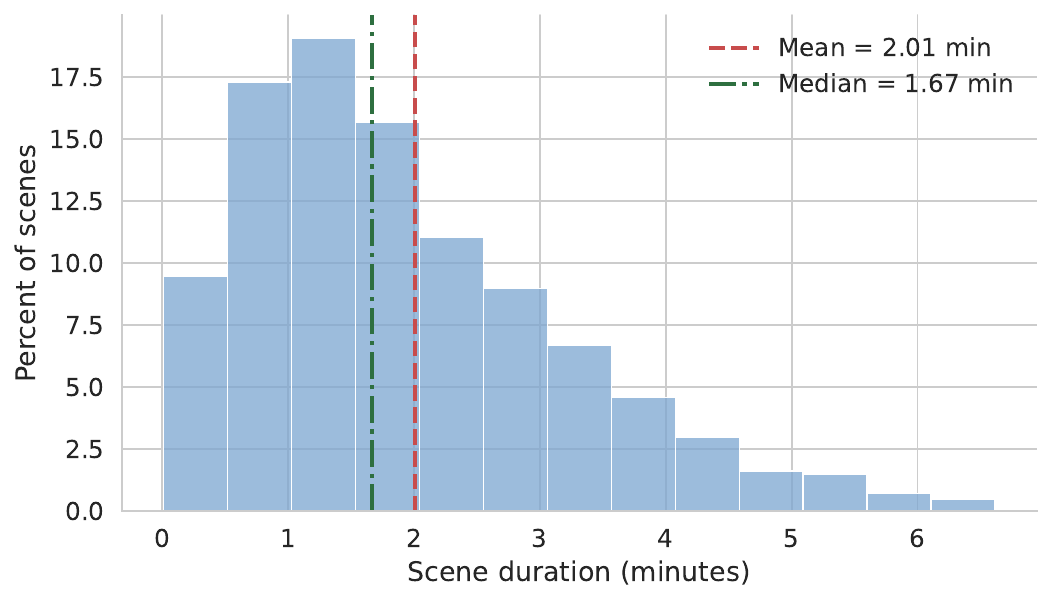}
        (a)\par\vspace{2pt}
    \end{minipage}
    \hfill
    \begin{minipage}[t]{0.48\linewidth}
        \centering
        \includegraphics[width=\linewidth]{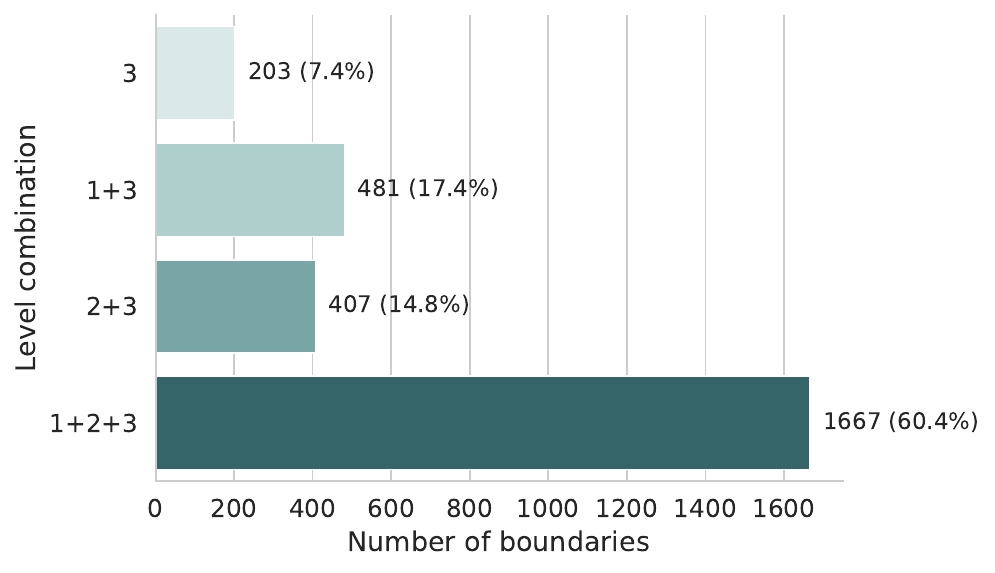}
        (b)\par\vspace{2pt}
    \end{minipage}
    \caption{Statistical overview of the MF$^2$-aligned portion of \ourdataset. The distribution of scene durations (a) shows that the dataset spans a broad range of narrative unit lengths, with a mean of 1.97 minutes and a median of 1.65 minutes. The distribution of scene-boundary annotations across cognitive-level combinations (b) shows that the dominant boundary types include Level~3 semantics, confirming that \ourdataset emphasizes semantic scene transitions beyond low-level visual discontinuities. Zoom in for details.}
    \label{fig:dataset_statistics}
\end{figure}

This section provides detailed statistics of the MF$^2$-aligned portion of \ourdataset that complement the summary in Section~\ref{sec:building_dataset}. We report the scale of the annotated movie collection and detailed counts of scene-boundary labels across cognitive-level combinations. Figure~\ref{fig:dataset_statistics} additionally visualizes the distributions of scene durations and boundary-label combinations, making the overall composition of the dataset easier to inspect.

Table~\ref{tab:appendix_dataset_stats} summarizes the exact statistics of the MF$^2$-aligned subset used in this appendix, including the number of movies, total duration, number of annotated boundaries, number of resulting scenes, and scene-duration statistics. Table~\ref{tab:appendix_boundary_stats} further reports detailed counts of scene-boundary labels, while Figure~\ref{fig:dataset_statistics}(b) provides the corresponding visual breakdown. Consistent with both the table and the bar chart, every annotated boundary includes Level~3 semantics, with a 100\% Level~3 presence rate, confirming that semantic scene transitions are the dominant annotation target in this subset. L1 and L2 frequently co-occur (Jaccard 0.587, $P(\text{L2}\mid\text{L1})=0.721$, $P(\text{L1}\mid\text{L2})=0.759$), but neither subsumes the other.

\section{Additional Qualitative Examples}
\label{app:qual_results}

\begin{figure}[t]
    \centering
    \includegraphics[width=\linewidth]{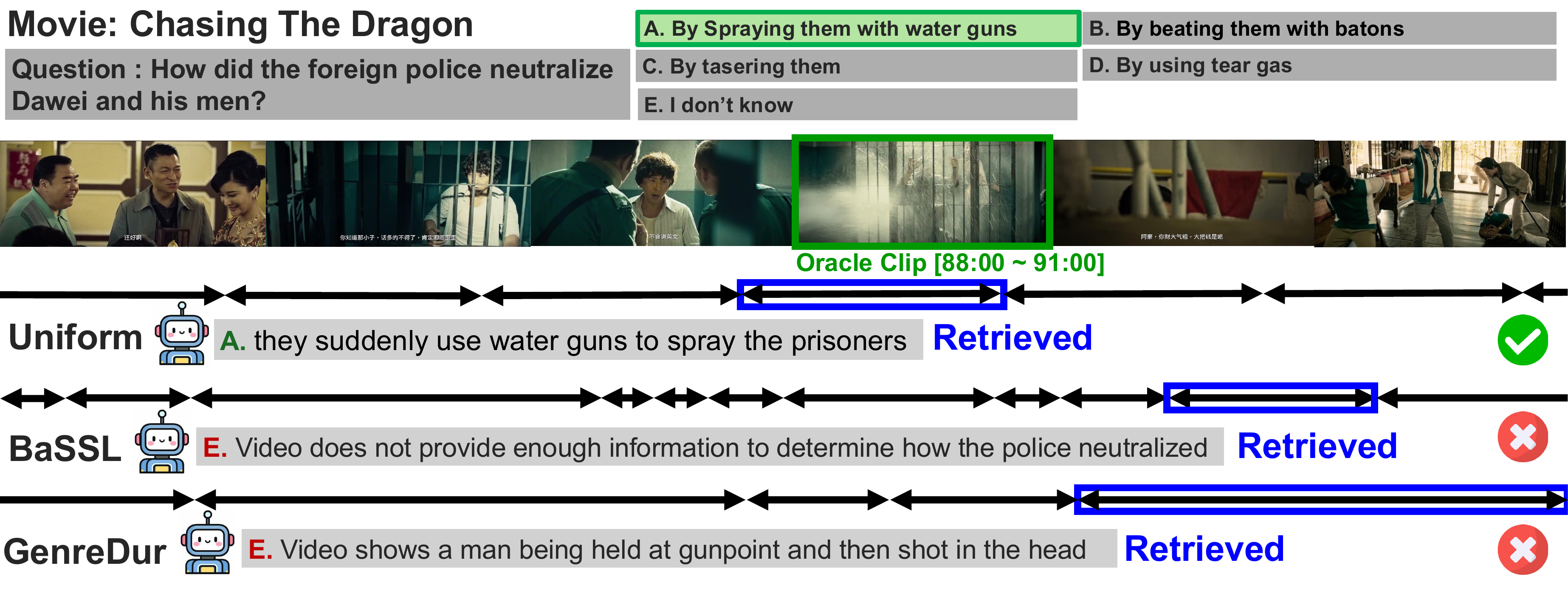}
    \caption{\textbf{Qualitative example on MovieStory101 Movie QA.} We compare retrieval results produced by different chunking strategies on a multiple-choice movie question from \textit{Chasing The Dragon}. The figure shows the oracle evidence clip together with the top retrieved chunk from each method and the corresponding QA prediction. Uniform chunking retrieves a clip that better matches the annotated evidence, whereas scene-based baselines may retrieve distractor clips that lead to incorrect answers or fallback predictions.}
    \label{fig:moviestory101_qual}
\end{figure}

\begin{figure}[t]
    \centering
    \includegraphics[width=\linewidth]{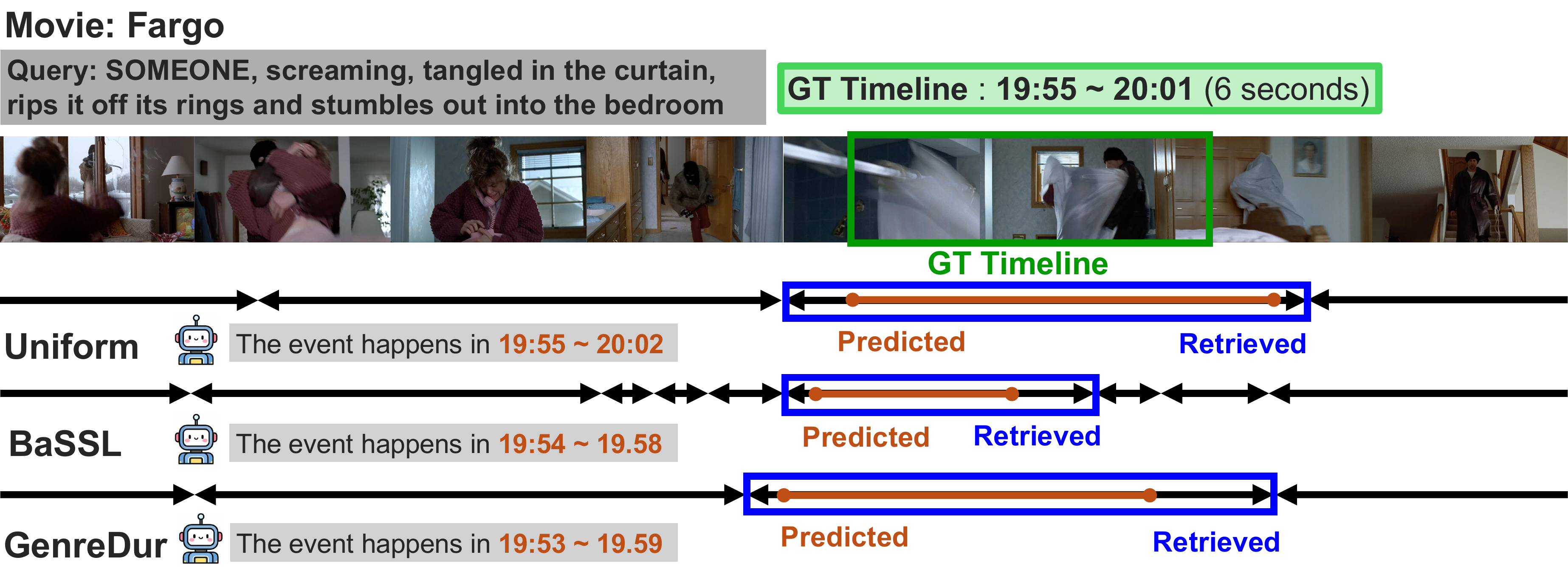}
    \caption{\textbf{Qualitative example on MAD Movie Temporal Grounding.} We compare retrieval and temporal grounding outputs under different chunking strategies for a query from \textit{Fargo}. The figure shows the ground-truth timeline together with the retrieved chunk and predicted temporal span from each method. Uniform chunking retrieves a candidate that more tightly covers the target moment, while scene-based baselines produce less accurate candidates and correspondingly weaker temporal localization.}
    \label{fig:mad_qual}
\end{figure}

We provide additional qualitative examples on MovieStory101 and MAD in Figure~\ref{fig:moviestory101_qual} and~\ref{fig:mad_qual}  to illustrate how chunk quality affects retrieval and downstream prediction. In both cases, errors in the retrieved chunk propagate directly to answer prediction or temporal localization.

\section{Narrative-Prompted VLM Baseline}
\label{app:zeroshot}
\ourbaseline prompts Qwen3.5-9B with our three-level definition, without any training or \ourdataset labels. Each movie is split into consecutive 180-second clips, and each clip is given to the model as video sampled at 1 FPS together with its ASR transcript. Predicted scenes are shifted to the global timeline, and the two scenes meeting at each clip cut are merged so that the cut itself does not create a boundary. Decoding is greedy, with up to two resamples (temperature 0.7, top-$p$ 0.8, top-$k$ 20) when the output is not a valid partition of the clip. The full prompt is shown in Figure~\ref{fig:zeroshot_prompt}.

\begin{figure}[h]
    \centering
    \begin{tcolorbox}[title={Zero-Shot Narrative Scene Segmentation Prompt}, colback=gray!2, colframe=black!30, arc=1mm, boxrule=0.4pt, width=0.95\textwidth, fonttitle=\bfseries, fontupper=\small]
    You are an expert video analyst and narrative researcher. Your task is to perform scene segmentation on the provided video data/script based on a strict multi-level criteria system.

    \textbf{Segmentation Criteria}\\
    You must evaluate and define scene boundaries based on the following three levels.\\
    $\bullet$ \textbf{Level 1 (Physical):} Visual or physical transitions in location or time.\\
    $\bullet$ \textbf{Level 2 (Character):} Transitions in the composition of the main characters.\\
    $\bullet$ \textbf{Level 3 (Narrative) [CRITICAL]:} Points where the OVERARCHING behavioral goal or primary conversation topic changes. Your scene segmentation MUST explicitly capture and include boundaries where a Level 3 transition occurs.

    \textbf{Anti-Over-segmentation Rule (Macro over Micro)}\\
    You must prevent over-segmentation.\\
    $\bullet$ \textbf{Focus on Macro-goals:} Ignore micro-level tactical shifts, brief interruptions, momentary distractions, or minor conversational pivots.\\
    $\bullet$ \textbf{Group short beats:} Combine smaller actions or dialogue exchanges into a single larger scene if they serve the same overarching narrative objective.

    \textbf{Semantic Sequence Rule}\\
    When encountering cross-cutting sequences where space and time alternate, but the shots share the exact same narrative goal and sound bridge, you must treat them as a \textbf{single scene}.

    \textbf{Output Format}\\
    You must output the final segmentation strictly as a JSON array of objects. Do not include any text outside the JSON.\\
    \texttt{[\{"start\_str": "00:00:00", "start\_sec": 0, "end\_str": "00:01:11", "end\_sec": 71, "levels": [1, 3]\},}\\
    \texttt{\ \{"start\_str": "00:01:11", "start\_sec": 71, "end\_str": "00:02:18", "end\_sec": 138, "levels": [2, 3]\}]}

    \textbf{Input Video Transcript (Dialogue \& Timestamps)}\\
    The following is the extracted dialogue from the video, with start and end timestamps in seconds. Correlate this dialogue with the visual changes in the video to determine narrative boundaries (Level 3).

    \texttt{\{transcript\_text\}}

    Analyze the video and the above transcript based on the Anti-Over-segmentation Rule, then provide ONLY the JSON output.
    \end{tcolorbox}
    \caption{Prompt used by \ourbaseline. One prompt is issued per 180-second clip, and the placeholder \texttt{\{transcript\_text\}} is filled with the timestamped ASR transcript of the clip.}
    \label{fig:zeroshot_prompt}
\end{figure}

\section{MAD-10 Validation Set}
\label{app:mad10}
We additionally annotate 10 movies from MAD~\citep{soldan2022mad}, selected by genre-stratified sampling before observing any downstream results. The annotation follows the same three-stage protocol as \ourdataset, yielding 603 scenes, and annotators had no access to the MAD queries or their ground-truth timestamps.

\section{Limitations and Social Impacts}
\label{app:limits_impacts}
\paragraph{Limitations.} Although \ourdataset covers full-length movies with substantial variation, it does not capture the full diversity of narrative cinema, including differences in genre, editing conventions, and production context. Our evaluation is also limited to full-length narrative films. Other formats may require adapting the operational definitions of the three levels, such as episodic television, where boundaries must reflect both local events and multi-episode arcs, or documentaries and talk shows, where coherent units are organized around topic, activity, or interaction, and we leave such cross-format evaluation to future work. Future work can expand this benchmark and develop automatic models that better recover narrative scene boundaries at scale.

\paragraph{Social impact.}
Our work may benefit long-form video retrieval, accessibility, and media analysis by providing context units that better reflect narrative events. Potential risks include misuse for large-scale indexing or monitoring of audiovisual content, as well as overgeneralizing one annotation scheme as a universal model of narrative structure. Because scene interpretation can vary across cultures, genres, and editing styles, we present \ourdataset as a research benchmark rather than a definitive account of how movies should be segmented.

\newpage